\documentclass{article}
\usepackage[preprint]{icml2026}
\usepackage{dblfloatfix}
\usepackage{placeins}
\usepackage{hyperref}
\usepackage{url}
\usepackage{booktabs}
\usepackage{amsfonts}
\usepackage{amsmath}
\usepackage{amssymb}
\usepackage{nicefrac}
\usepackage{microtype}
\usepackage{xcolor}
\usepackage{graphicx}
\usepackage{multirow}
\usepackage{pifont}
\usepackage[most]{tcolorbox}
\usepackage{subcaption}

\usepackage[table]{xcolor}

\newcommand{\eventsub}[1]{\vspace{0.4em}\noindent$\triangleright$ \textit{#1:}\;}   

\definecolor{SectionGray}{RGB}{238,238,238}
\definecolor{LightGreen}{RGB}{222,245,224}
\newcommand{\cmark}{\ding{51}}
\newcommand{\nmark}{\ding{55}}

\icmltitlerunning{OmniVL-Guard Pro}

\newlength{\fulltextwidth}
\begin{document}

\twocolumn[
  \icmltitle{OmniVL-Guard Pro: A Tool-Augmented Agent for Omnibus Vision-Language Forensics}

  \icmlsetsymbol{equal}{*}
  \icmlsetsymbol{cor}{$\dagger$}

  \begin{icmlauthorlist}
    \icmlauthor{Jinjie Shen}{equal,hfut,whu,lion}
    \icmlauthor{Zheng Huang}{equal,whu}
    \icmlauthor{Yuchen Zhang}{xjtu}
    \icmlauthor{Yujiao Wu}{csiro} \\
    \icmlauthor{Yaxiong Wang}{cor,hfut,lion}
    \icmlauthor{Lechao Cheng}{hfut,lion}
    \icmlauthor{Shengeng Tang}{hfut}
    \icmlauthor{Tianrui Hui}{hfut,lion}
    \icmlauthor{Nan Pu}{hfut,lion}
    \icmlauthor{Zhun Zhong}{cor,hfut,lion}
  \end{icmlauthorlist}

  \begin{center}
    \vspace{0.05in}
    {\scriptsize $^{1}$School of Computer Science and Information Engineering, Hefei University of Technology, Hefei, China\par}
    \vspace{0.04in}
    {\scriptsize $^{2}$Wuhan University, Wuhan, China \quad $^{3}$Lab for Intelligence and visiON (LION)\par}
    \vspace{0.04in}
    {\scriptsize $^{4}$Xi'an Jiaotong University \quad $^{5}$CSIRO\par}
    \vspace{0.06in}
    {\scriptsize $^{*}$Equal contribution. \quad $^{\dagger}$Corresponding author: \texttt{wangyx@hfut.edu.cn}; \texttt{zhunzhong007@gmail.com}}
  \end{center}

  \icmlaffiliation{hfut}{School of Computer Science and Information Engineering, Hefei University of Technology, Hefei, China}
  \icmlaffiliation{whu}{Wuhan University, Wuhan, China}
  \icmlaffiliation{lion}{Lab for Intelligence and visiON (LION)}
  \icmlaffiliation{xjtu}{Xi'an Jiaotong University}
  \icmlaffiliation{csiro}{CSIRO}

  \icmlcorrespondingauthor{Yaxiong Wang}{wangyx@hfut.edu.cn}
  \icmlcorrespondingauthor{Zhun Zhong}{zhunzhong007@gmail.com}

  \icmlkeywords{Forgery Detection, Vision-Language Model, Reinforcement Learning, Tool-Augmented Agent}

  \vskip 0.03in
  \centerline{\includegraphics[width=0.94\textwidth]{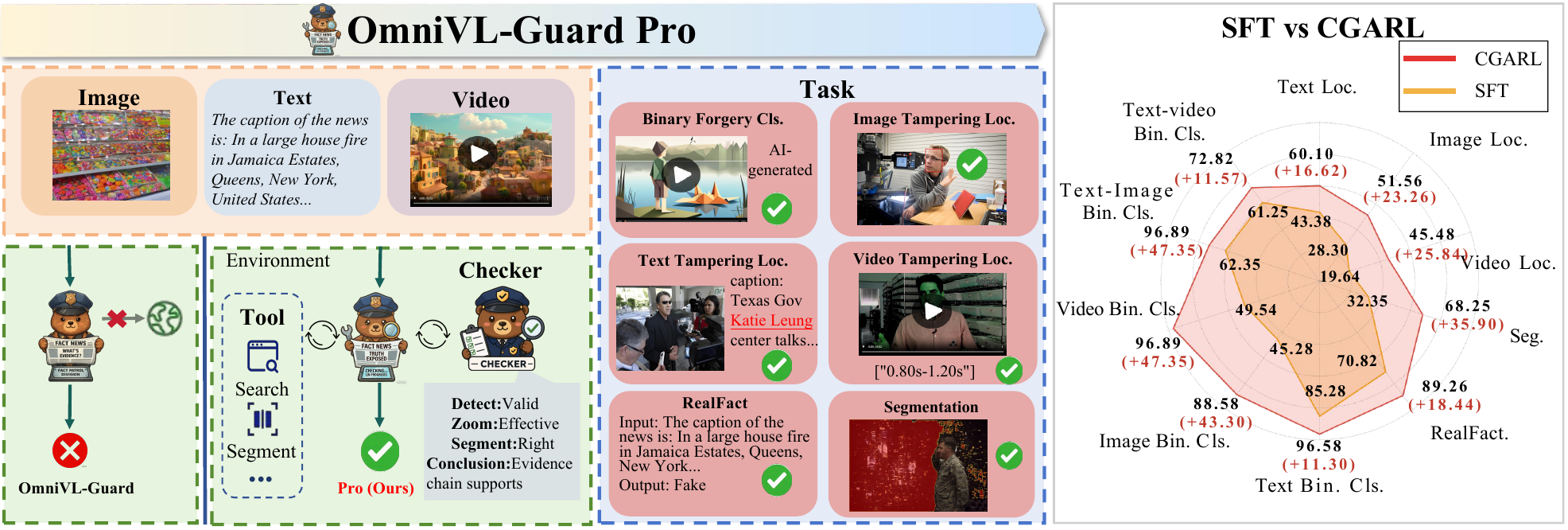}}
  \vskip 0.04in
  \begin{center}
    \begin{minipage}{0.95\textwidth}
      \vspace{-0.3cm}
      \captionof{figure}{This work tackles omnibus vision-language forgery detection, grounding, and segmentation (left). In this unified setting, models relying on parametric memory alone cannot handle manipulations that require external clues or fine-grained inspection. In response, we propose OmniVL-Guard Pro, a tool-augmented agent trained with Checker-Guided Agentic RL (CGARL), achieving evidence-grounded reasoning and consistent gains over SFT across all tasks (right).}
      \label{fig:placeholder}
    \end{minipage}
  \end{center}

  \vskip 0.08in
  \begin{abstract}
Existing vision-language forgery detection and grounding methods operate under a closed-world paradigm, assuming verification can be completed by the model alone. However, self-contained MLLMs are constrained by finite parametric knowledge, static training corpora, and limited perceptual resolution, creating a practical ceiling in dynamic open-world forensics---particularly for real-time event verification requiring external clues and forgery segmentation demanding fine-grained scrutiny of local manipulations. To address these limitations, we shift from scaling up the self-contained model toward reaching beyond it. We propose \textbf{OmniVL-Guard Pro}, a tool-augmented agent that extends unified forensics from closed-world prediction to open-world clues-driven reasoning. OmniVL-Guard Pro integrates a tool environment spanning real-time event search, local cropping and zooming, edge-anomaly screening, face detection, video frame extraction, and SAM3-based segmentation. To generate high-quality tool-reasoning trajectories, we introduce \textbf{Tree-Structured Self-Evolving Tool Trajectory Generation}, which produces diverse trajectories through seed guidance, guider-free self-evolution, and weakly-hinted hard sample synthesis, yielding the Full-Spectrum Tool Reasoning (FSTR) dataset for training. We further propose \textbf{Checker-Guided Agentic Reinforcement Learning} (CGARL), which provides process-level supervision to penalize cases where the answer is correct but the reasoning is distorted. Extensive experiments demonstrate that OmniVL-Guard Pro achieves state-of-the-art performance across various tasks, and exhibits strong zero-shot generalization. The FSTR dataset and code for OmniVL-Guard Pro will be publicly released at \url{https://github.com/shen8424/OmniVL-Guard-Pro}.
  \end{abstract}

  \vskip 0.1in
]

\section{Introduction}
\label{sec:introduction}

The rapid evolution of generative AI has significantly lowered the barrier for creating manipulated content across diverse modalities to infiltrate real-world social media ecosystems at an unprecedented pace~\cite{intro1,intro2,intro3}. Consequently, forgery detection, manipulation localization, and forgery segmentation across primary modalities including images, videos, and text have garnered substantial attention~\cite{samm, fkaowl}. 
Despite recent advancements such as OmniVL-Guard~\cite{omnivl-guard}, which establishes a unified framework for omnibus vision-language forgery detection and grounding, 
existing methods still mainly operate under a \emph{closed-world} paradigm, which implicitly assumes that \textbf{\emph{the verification process can be completed by the model itself.}}

However, self-contained MLLMs~\cite{deepseek-r1,amd} remain constrained by finite parameters, static training corpora, and limited native perceptual resolution, which impose a practical ceiling in dynamic open-world forensics. This limitation is evident in two representative settings: Real-time Event Verification requires up-to-date external clues beyond parametric memory, while Forgery Segmentation demands fine-grained scrutiny of local boundaries, texture discontinuities, facial details, and semantic inconsistencies for pixel-level localization. Although scaling model size or training data could partially enhance these abilities, such a ``scaling-up'' strategy is costly and insufficient for continuously evolving real-world forgeries. We therefore shift from merely scaling up the self-contained model to reaching out beyond it, enabling the model to retrieve real-world knowledge and invoke specialized perception tools when necessary. In this sense, these two tasks jointly motivate an open-world forensic agent that \textbf{\emph{dynamically acquires external knowledge and fine-grained visual artifacts to transcend the intrinsic limits of the base model}}.

To this end, we first construct a tool-enhanced environment that covering key capabilities such as real-time event search, local cropping, and segmentation (Sec.~\ref{sec:tool_env}). This environment enables the model to evolve from a self-contained predictor into an clues-seeking agent that actively gathers, examines, and
integrates external clues.

With the tool environment as the base, the agent needs to decide when and which tools is invoked. Cultivating this ability requires high-quality tool-based reasoning trajectory data.
Unlike pure-text Chain-of-Thought (CoT), tool-based reasoning trajectory simultaneously involves tool selection, parameter specification, observation understanding, and stopping decisions, making it substantially more difficult to construct. Even state-of-the-art MLLMs~\cite{gpt,gemini,seedvl} still exhibit limited exploration quality in real tool environments, and thus cannot directly produce sufficiently high-quality training trajectories. Conversely, directly injecting GT or answer-aware cues would cause the model to organize tool usage around the known answer, thereby introducing hindsight bias~\cite{omnivl-guard}. To address this challenge, we propose a \textbf{Tree-Structured Self-Evolving Forensic Tool Trajectory Generation} framework. In the seed stage, it efficiently constructs high-fidelity trajectories through tree expansion and branch pruning. It then gradually scales up the data through unguided self-evolution, while further strengthening long-tail hard samples with weak hints, thereby providing cold-start data for subsequent tool-augmented RL.

Taking the trajectory data, we finally train the model for detection, grounding and segmentation with the aid of external tools. 
However, merely teaching the model to invoke tools is still insufficient. The core of forgery detection, grounding, and segmentation lies not only in obtaining external clues/artifacts, but also in forming correct and consistent judgments based on such evidence. The model still exhibits a pseudo-success pattern where the \textbf{\emph{``answer is accidentally correct but the process is distorted''}}: although the final answer happens to be correct, the intermediate tool calls, observational evidence, and final judgment do not form a consistent and verifiable support relation, and such deviations are difficult for the model itself to identify. To tackle this problem, we further propose \textbf{Checker-Guided Agentic Reinforcement Learning} (CGARL). Specifically, we introduce a \emph{Checker} to build a Multi-Agent system, which provides process-level supervision over intermediate reasoning, tool usage, and observational evidence, explicitly assessing whether they jointly support the final judgment. Through this mechanism, the model is encouraged to genuinely organize its reasoning around evidence, rather than remaining at the stage of merely knowing how to call tools without properly using the acquired evidence.

Built upon the above designs, we propose \textbf{OmniVL-Guard Pro}, an open-world tool-augmented Agent model. By integrating Tree-Structured Self-Evolving Forensic Tool Trajectory Generation with Checker-guided process-supervised RL, OmniVL-Guard Pro advances OmniVL-Guard from a closed-world unified forensic model into an open-world evidence-driven intelligent agent. Overall, our main contributions are summarized as follows:

(1) \textbf{Open world Agent for Vision-Language Forensics.} We propose OmniVL-Guard Pro, a tool-augmented Agent that performs forgery detection, grounding, and segmentation with external support across text, image, and video modalities, supporting  real-time event verification and fine-grained forgery segmentation. We also construct a benchmark for real-time event verification.

(2) \textbf{FSTR Dataset.} We construct \emph{Full-Spectrum Tool Reasoning} dataset, consisting of $\mathrm{FSTR}_{\mathrm{sft}}$ for cold-start SFT and $\mathrm{FSTR}_{\mathrm{rl}}$ for subsequent tool-augmented RL. $\mathrm{FSTR}_{\mathrm{sft}}$ contains high-fidelity tool-search trajectories, while $\mathrm{FSTR}_{\mathrm{rl}}$ supports subsequent tool-augmented RL.

(3) \textbf{Checker-Guided Agentic RL.} We propose Checker-Guided Agentic RL, which introduces process-level supervision beyond answer-level rewards to encourage evidence-consistent reasoning. It mitigates the pseudo-success pattern where the ``answer is correct but the process is distorted,'' improving both Agent performance and reasoning-evidence consistency.

\vspace{-0.3cm}
\section{Tool-Enhanced Environment}
\vspace{-0.2cm}
\label{sec:open_world_env}

\label{sec:tool_env}

To enable open-world evidence acquisition beyond intrinsic model capabilities, we construct a tool environment for unified vision-language forgery detection, grounding, and segmentation.

\textbf{Real-time event search tool.}
This tool searches the web given a keyword query (e.g., time, person, location), returning the most relevant results and supporting clues for real-time fact verification. 

\textbf{Local cropping tool.}
Given normalized coordinates, this tool extracts the specified region as a new image context for further inspection.

\textbf{Local zooming tool.}
Given a candidate region, this tool returns an enlarged view for fine-grained inspection of boundary discontinuities, texture anomalies, local blur, and subtle artifacts.

\textbf{Edge-anomaly screening tool} scans the image using local edge density, Laplacian energy, and neighborhood discrepancy, and returns candidate regions for cropping and zooming.

\textbf{Face detection tool} returns face bounding boxes based on InsightFace~\cite{insightface}, supporting facial region inspection in portrait and person-centric media.

\textbf{Video clip frame extraction tool.}
Given a candidate temporal segment, this tool extracts frames for further examination using the image-level tools above.

\textbf{SAM3 segmentation tool} runs SAM3~\cite{sam3} to produce a pixel-level mask from model-derived prompts: bounding box, positive points inside the target, and negative points in the background.

\section{Tree-Structured Self-Evolving Tool Trajectory Generation}
\label{sec:tool_traj_gen}

\begin{figure*}[tb]
    \centering
    \includegraphics[width=\fulltextwidth]{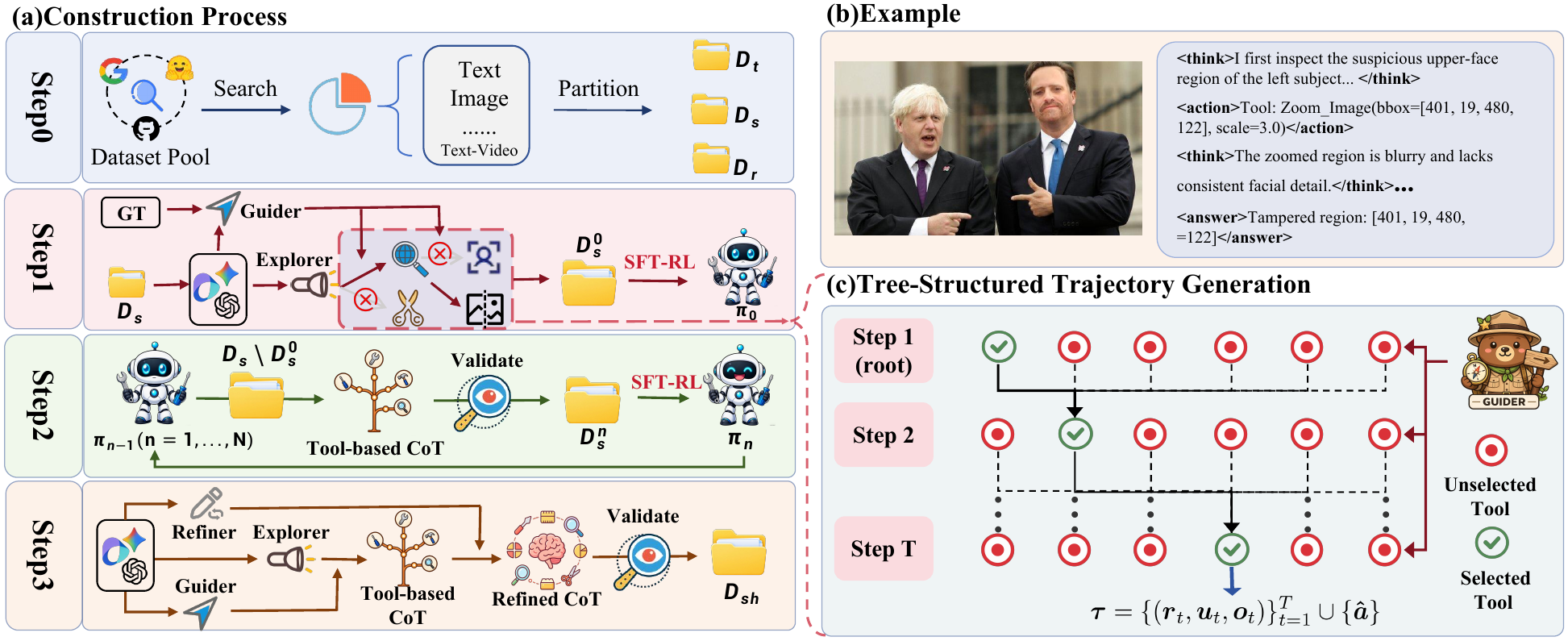}
    \caption{(a)The construction of FSTR$_\text{sft}$. (b) An example from FSTR$_\text{sft}$. (c) The process of Tree-Structured Trajectory Generation.}
    \label{fig:dataset}
    \vspace{-0.6cm}
\end{figure*}

High-quality tool-search trajectories are a critical prerequisite for the cold-start phase of tool-augmented reinforcement learning.  However, constructing such data in unified vision-language scenarios faces a new \textbf{efficiency--bias dilemma}: on the one hand, if we directly rely on general-purpose MLLMs to freely explore in real tool environments, they often produce a large number of low-quality trajectories since stable tool-use capabilities have not yet been established; on the other hand, if GT or answer-aware cues are injected to force-guide the search process, hindsight bias will be introduced, causing the model to learn answer-conditioned tool invocation patterns rather than transferable evidence-seeking capabilities. To address this issue, we propose a \textbf{Tree-Structured Self-Evolving Tool Trajectory Generation} strategy. Overall, this strategy consists of four stages, as shown in Figure~\ref{fig:dataset}.

\vspace{-0.3cm}
\subsection{Source Data Collection}
\vspace{-0.1cm}
\label{sec:in-domain-datasets}
We first collect raw data from a series of existing datasets to cover the major modalities and task types involved in unified vision-language forensics. Specifically, for the uni-modal text domain, we adopt FakeNewsCorpus~\cite{fakenewsnet} and MCFEND~\cite{mcfend}; for the single-image modality, we select FakeClue~\cite{fake-vlm}, LOKI~\cite{loki}, ForgeryNet~\cite{forgerynet}, OpenSDI~\cite{opensdi}, and AutoSplice~\cite{autosplice}; for the video modality, we use GenVideo~\cite{genvideo}, DVF~\cite{dvf}, and ForgeryNet~\cite{forgerynet}; for the image--text modality, we adopt SAMM~\cite{samm}, MDSM~\cite{amd}, DGM$^4$~\cite{hammer}, and NewsCLIPpings~\cite{newsclippings}; and for the text--video modality, we select FakeSV~\cite{fakesv}. These datasets jointly cover diverse tasks related to forgery detection, manipulation grounding, and semantic segmentation. For the real-time event verification task, we use RealFact constructed in the Appendix~\ref{sec:realtime_event_benchmark}.
Finally, we partition all source data into three parts: $\mathcal{D}_s$ for SFT cold-start data construction, $\mathcal{D}_r$ for subsequent RL data construction, and $\mathcal{D}_t$ for in-domain testing. The selection of out-of-domain zero-shot test data is detailed in Sec.~\ref{sec:experiments}.

\vspace{-0.3cm}
\subsection{Tool Trajectory Seed Guidance}
\vspace{-0.1cm}
\label{sec:seed_traj_guidance}

On a small-scale subset of $\mathcal{D}_s$, we first construct a high-fidelity seed set of tool-search trajectories. To this end, we define an MLLM expert pool
\begin{equation}
\mathcal{M}=\{\text{GPT5.4},\ \text{Gemini3.1},\ \text{Seed2.0}\}.
\end{equation}
For each sample, we randomly select two distinct models from $\mathcal{M}$ to serve as the \textit{Explorer} and the \textit{Guider}, respectively. The \textit{Explorer} is responsible for executing search in the real tool environment, while the \textit{Guider}, with the final answer $a^\star$ injected, determines the validity of the tool attempt performed at the current step. Based on this design, seed trajectory generation is not a linear rollout along a single path, but rather a tree-structured search process composed of branch expansion and branch pruning, as shown in Figure~\ref{fig:dataset}(c).

Formally, a tool-search trajectory is denoted as
\begin{equation}
\label{eq2}
\tau = \{(r_t, u_t, o_t)\}_{t=1}^{T} \cup \{\hat{a}\},
\end{equation}
where $r_t$ denotes the model's reasoning process at step $t$, which is formed based on previous tool-returned observations and determines the next tool to invoke as well as its parameters. $u_t$ denotes the selected tool and its parameters at the current step, $o_t$ denotes the observation returned by the tool, and $\hat{a}$ denotes the final prediction produced by the model at the end of the trajectory.

At step $t$, given the current history 
$\{(r_i,u_i,o_i)\}_{i=1}^{t-1}$, the \textit{Explorer} expands the current node by sequentially attempting each tool in the tool set $\mathcal{T}$ (Sec.~\ref{sec:tool_env}). For each candidate tool, the \textit{Explorer} first generates the reasoning process $r_t$, which analyzes the previous observations and specifies both the tool to be invoked and its parameters, such as a search query, a crop box, or a zooming region. The corresponding tool invocation is denoted as $u_t$, and is executed in the real tool environment to obtain the returned observation $o_t$. The \textit{Guider}, with the final answer $a^\star$ injected, then evaluates whether this candidate step $(r_t,u_t,o_t)$ is helpful for solving the final task. If it is judged invalid, the current branch is immediately terminated, and the \textit{Explorer} turns to the next candidate tool under the same history state; if it is judged valid, $(r_t,u_t,o_t)$ is appended to the trajectory, and the resulting new state is further expanded in the next step following the same procedure.


After the \textit{Explorer} completes the search and outputs the final prediction $\hat{a}$, we further conduct quality checking on the trajectory, requiring the final answer to be correct,  the overall trajectory to be coherent, the intermediate reasoning and tool usage to support the final conclusion, and the trajectory to contain no obvious ineffective steps. The detailed criteria for the \textit{Guider} judgment and the subsequent quality checking process are provided in Appendix~\ref{sec:guider_criteria}~\ref{sec:quality_checking}. All trajectories that pass the checking process are included in the seed set $\mathcal{D}_s^0$. Subsequently, we perform SFT on Qwen3VL-8B~\cite{qwen3vl} using $\mathcal{D}_s^0$, followed by RL on $\bar{\mathcal{D}}_r \subset \mathcal{D}_r$, yielding the initial policy $\pi_0$.

\subsection{Guider-free Self-Evolving Expansion}
\label{sec:self-evolving}

After obtaining the initial policy $\pi_0$, we further design a Guider-free self-evolving expansion mechanism to continuously scale up high-quality tool-augmented reasoning trajectories.
Specifically, in the $n$-th iteration, we use $\pi_{n-1}$ to perform multi-turn rollout reasoning on $\mathcal{D}_s \setminus \mathcal{D}_s^0$, producing candidate tool trajectories. Different from the seed stage, we remove the answer-aware Guider from the rollout process, so the current policy autonomously explores the tool environment without Guider-based step-level judgment or branch pruning. For each candidate trajectory, we apply the same checking and filtering procedure as in Sec.~\ref{sec:seed_traj_guidance}, retaining only samples whose final answers are correct, whose trajectories are globally self-consistent, whose tool usage and reasoning processes support the final conclusion, and which contain no obvious ineffective steps (Appendix~\ref{sec:quality_checking}). All trajectories that pass the filtering process are combined with $\mathcal{D}_s^0$ to form the current-round dataset $\mathcal{D}_s^n$. We then re-initialize SFT from Qwen3VL-8B using $\mathcal{D}_s^n$, followed by RL on $\bar{\mathcal{D}}_r$, obtaining the $n$-th round policy $\pi_n$. We repeat this self-evolving process for $N=4$.

\subsection{Weakly-Hinted Hard Trajectory Synthesis}

Although self-evolution can substantially improve the coverage of valid trajectories, standard rollout could still fail to produce any accepted trajectory for samples whose correct tool-use paths require sparse clues, precise local regions, or non-obvious search keywords. These persistently failed samples constitute the long-tail portion of the trajectory distribution. To complement the self-evolved data on such hard cases, we further design a weakly-hinted hard trajectory synthesis mechanism.

Specifically, for samples that consistently fail during the self-evolution stage, following Sec.~\ref{sec:seed_traj_guidance}, we randomly select two models from $\mathcal{M}$ to serve as \textit{Explorer} and \textit{Guider}, respectively. The \textit{Explorer} sequentially attempts each tool in the tool set $\mathcal{T}$ at each step to generate candidate trajectories, while the \textit{Guider}, with the final answer $a^\star$ injected, judges the tool attempt at the current step. Different from Sec.~\ref{sec:seed_traj_guidance}, the \textit{Guider} not only determines whether the current tool selection is valid, but also provides usage hints for the tool, namely necessary parameters or local clues related to the current tool invocation, to narrow down the search space at the current step for the \textit{Explorer}. For example, in image scenarios, the \textit{Guider} provides a rough candidate region to assist cropping or zooming. Based on these weak hints, the \textit{Explorer} performs tool search and generates candidate trajectories.

After obtaining the weakly-hinted candidate trajectories, we further randomly select another model from $\mathcal{M}$ to serve as the \textit{Refiner}, which rewrites the reasoning process $r_t$ at each step of the trajectory to eliminate explicit traces introduced by weak hints, preventing such answer-aware hints from contaminating the trajectory and inducing hint-dependent reasoning. We then apply the same quality checking procedure as in Sec.~\ref{sec:seed_traj_guidance}.
All trajectories that pass the checking process are included in the hard-sample set $\mathcal{D}_{sh}$. The rewriting details are provided in Appendix~\ref{sec:refiner}.

\subsection{Dataset Statistics}
\label{sec:dataset_statistics}
Through the aforementioned process, we obtain the Full-Spectrum Tool Reasoning (FSTR) dataset. It consists of two parts: the tool-search trajectory set for cold-start SFT, $\mathrm{FSTR}_{\mathrm{sft}}=\mathcal{D}_s^N \cup \mathcal{D}_{sh}$; and the dataset for RL training, $\mathrm{FSTR}_{\mathrm{rl}}=\mathcal{D}_r \setminus \bar{\mathcal{D}}_r$. $\mathrm{FSTR}_{\mathrm{sft}}$ contains 69,323 samples, where $\mathcal{D}_{sh}$ accounts for approximately 17\% and $\mathcal{D}_s^0$ accounts for approximately 8\%; $\mathrm{FSTR}_{\mathrm{rl}}$ contains 81,087 samples. FSTR covers five core modalities, including text, image, video, text-image, and text-video, and supports full-spectrum vision-language classification tasks, omni-modal grounding tasks, image-modal segmentation tasks, and real-time event verification tasks.

\section{Checker-Guided Agentic Reinforcement Learning}
\label{sec:cgarl}

After obtaining the cold-start tool trajectory data, we adopt a three-stage training strategy. Specifically, we first perform SFT on $\mathrm{FSTR}_{\mathrm{sft}}$ and then split $\mathrm{FSTR}_{\mathrm{rl}}$ into two disjoint subsets, $\mathcal{D}_{\mathrm{rl}}^{1}$ and $\mathcal{D}_{\mathrm{rl}}^{2}$. In the first stage, we conduct outcome-level RL on $\mathcal{D}_{\mathrm{rl}}^{1}$, yielding a tool-capable policy $\pi_{\mathrm{ans}}$. In the second stage, we rollout $\pi_{\mathrm{ans}}$ on $\mathcal{D}_{\mathrm{rl}}^{2}$ to construct process supervision data and train the process discrimination model $\pi_{\mathrm{checker}}$. The third stage further performs Checker-guided process-supervised RL on $\mathcal{D}_{\mathrm{rl}}^{2}$, jointly improving answer correctness, process consistency, and tool-use efficiency. The rationale behind this design is that effective Checker supervision requires a policy with stable tool-use capabilities; otherwise, process-level signals would be noisy when the model still fails to perform reliable detection, grounding, or segmentation. 


\subsection{Stage I: Outcome-Driven RL with ARSPO}

We first perform SFT on Qwen3VL-8B using $\mathrm{FSTR}_{\mathrm{sft}}$, obtaining the initial policy $\pi_{\mathrm{sft}}$. For a sample with a tool trajectory, we write it as
\begin{equation}
\tau = \{(r_t, u_t, o_t)\}_{t=1}^{T} \cup \{\hat{a}\},
\end{equation}
where the definitions of $r_t$, $u_t$, $o_t$, and $\hat{a}$ are consistent with Eq.~\ref{eq2}. Since $o_t$ is returned by the environment rather than generated by the model, we do not compute loss over these tokens during SFT. Instead, the language modeling loss is only applied to model-generated tokens corresponding to $r_t$, $u_t$ and $\hat{a}$.

Let the complete token sequence of $\tau$, including all $T$ reasoning-tool interaction steps and the final answer, be $y=(y_1,\dots,y_{|y|})$, and define a token-level mask $m_i\in\{0,1\}$, where positions corresponding to tool-returned observations $o_t$ take $m_i=0$, and all other model-generated positions take $m_i=1$. The SFT loss is therefore defined as
\vspace{-0.2cm}
\begin{multline}
\mathcal{L}_{\mathrm{sft}}
= -\sum_{(x,\tau)\in \mathrm{FSTR}_{\mathrm{sft}}}
\frac{1}{\sum_{i=1}^{|y|} m_i} \\
\times \sum_{i=1}^{|y|}
m_i \log \pi_{\theta}(y_i\mid x,y_{<i}).
\end{multline}
In this way, the model learns how to generate reasonable tool invocations, parameter specifications, intermediate reasoning, and final answers, rather than fitting the contents returned by the tools.

Subsequently, we split $\mathrm{FSTR}_{\mathrm{rl}}$ into two parts, $\mathcal{D}_{\mathrm{rl}}^{1}$ and $\mathcal{D}_{\mathrm{rl}}^{2}$, where $\mathcal{D}_{\mathrm{rl}}^{1}$ is used for outcome-level RL, and $\mathcal{D}_{\mathrm{rl}}^{2}$ is reserved for subsequent Checker-guided process-supervised optimization.

For each task question $x$ in $\mathcal{D}_{\mathrm{rl}}^{1}$, the current policy $\pi_\theta$ performs inference in the real tool environment and produces a complete response$y=(\tau,\hat{a})$. Then we define the total reward as
\begin{equation}
R(y,x)=R_{\mathrm{ans}}(y,x)+R_{\mathrm{fmt}}(y)+R_{\mathrm{rep}}(y),
\end{equation}
where $R_{\mathrm{ans}}$, $R_{\mathrm{fmt}}$, and $R_{\mathrm{rep}}$ denote the answer reward, format reward, and repetition penalty, respectively. $R_{\mathrm{ans}}$ is computed by accuracy for classification tasks, and by overlap-based metrics such as IoU or Dice for grounding and segmentation tasks. Details are provided in Appendix~\ref{sec:reward_functions}.


We adopt ARSPO~\cite{omnivl-guard} to update the policy, whose optimization objective is formulated as follows:
\vspace{-0.2cm}
\begin{equation}
\begin{split}
\mathcal{J}_{\mathrm{arspo}}(\theta)
&= \sum_{k=1}^{K}\frac{|\mathcal{D}_{\mathrm{rl},k}^{1}|}{|\mathcal{D}_{\mathrm{rl}}^{1}|}
\; \mathbb{E}_{\substack{x\sim \mathcal{D}_{\mathrm{rl},k}^{1},\\ \{y_i\}\sim \pi_{\theta_{\mathrm{old}}}}} \\
&\qquad \times \Bigg[
\frac{l_{k,s}}{G}\sum_{i=1}^{G}
\frac{1}{|y_i|}
\sum_{t=1}^{|y_i|}
f_{i,t}\big(r_{i,t}(\theta)\big)\hat{A}_{i,k}
\Bigg],
\end{split}
\label{eq:arspo_tool}
\end{equation}
where $K$ denotes the number of task categories, $\mathcal{D}_{\mathrm{rl},k}^{1} \subset \mathcal{D}_{\mathrm{rl}}^{1}$ denotes the data subset corresponding to the $k$-th task, and $G$ denotes the number of sampled responses for each question. $y_i$ denotes the $i$-th complete response, and $|y_i|$ is its total token length. $\hat{A}_{i,k} = \frac{A_{i,k} - \mu}{\sigma}$ is the normalized advantage, with $\mu$ and $\sigma$ being the mean and standard deviation of the rewards $\{A_{i,k}\}_{i=1}^G$. $f_{i,t}(\cdot)$ is the token-level weighting function in the form of SAPO~\cite{sapo} (Detailed in Appendix~\ref{sec:advantage_function}), and $l_{k,s}$ denotes the dynamic coefficient for the $k$-th task at training step $s$. $r_{i,t}=\frac{\pi_{\theta}(y_{i,t}|q,y_{i,<t})}{\pi_{\theta_{\text{old}}}(y_{i,t}|q,y_{i,<t})}$ is the probability ratio.

Consistent with the SFT stage, the token-level optimization in Eq.~\eqref{eq:arspo_tool} is only applied to model-generated tokens. After this stage, the model is updated from $\pi_{\mathrm{sft}}$ to $\pi_{\mathrm{ans}}$, which serves as the initialization policy for subsequent Checker-guided process-supervised optimization.

\begin{table*}[tb]
\centering
\setlength{\tabcolsep}{2.8pt}
\caption{Performance comparison on the in-domain test set $\mathcal{D}_t$.}
\label{tab:in-domain}
\resizebox{\textwidth}{!}{
\begin{tabular}{l cccccc cccc}
\toprule
\multirow{2}{*}{\textbf{Method}} &
\multicolumn{6}{c}{\textbf{Binary Classification (ACC)}} &
\multirow{2}{*}{\textbf{Image Loc.}} &
\multirow{2}{*}{\textbf{Text Loc.}} &
\multirow{2}{*}{\textbf{Video Loc.}} &
\multirow{2}{*}{\textbf{Seg.}} \\
\cmidrule(lr){2-7}
& Text & Image & Video & Text-Image & Text-Video & RealFact & (IoU) & (F1) & (tIoU) & (Dice) \\
\midrule

\rowcolor{SectionGray} \multicolumn{11}{l}{\textit{Multi-modal Large Language Models}} \\
Qwen3VL-235B~\cite{qwen3vl} & 89.23 & 65.32 & 70.80 & 58.42 & 54.31 & 55.74 & 16.37 & 31.25 & 21.43 & ---  \\
Llama3.2-90B~\cite{llama3.2} & 83.47 & 60.34 & 63.28 & 55.91 & 49.84 & 52.36 & 14.85 & 25.99 & 19.54 & ---  \\
\midrule

\rowcolor{SectionGray} \multicolumn{11}{l}{\textit{Text-Image Detection Methods}} \\
HAMMER~\cite{hammer} & --- & --- & --- & 71.23 & --- & --- & 48.53 & 40.86 & --- & --- \\
FKA-Owl~\cite{fkaowl} & --- & --- & --- & 72.08 & --- & --- & --- & --- & --- & --- \\
AMD~\cite{amd} & --- & --- & --- & 65.34 & --- & --- & 35.91 & 37.64 & --- & --- \\
\midrule

\rowcolor{SectionGray} \multicolumn{11}{l}{\textit{Text Detection Methods}} \\
Bert~\cite{bert} & 61.45 & --- & --- & --- & --- & 51.37 & --- & 35.91 & --- & --- \\
LUKE~\cite{luke} & 63.72 & --- & --- & --- & --- & 50.95 & --- & 38.28 & --- & --- \\
\midrule

\rowcolor{SectionGray} \multicolumn{11}{l}{\textit{Image Detection Methods}} \\
D$^3$~\cite{D3} & --- & 88.46 & --- & --- & --- & --- & --- & --- & --- & --- \\
Fake-VLM~\cite{fake-vlm} & --- & 90.39 & --- & --- & --- & --- & --- & --- & --- & --- \\
\midrule

\rowcolor{SectionGray} \multicolumn{11}{l}{\textit{Video Detection Methods}} \\
FakeSV-VLM~\cite{fakesv-vlm} & --- & --- & 98.81 & --- & --- & --- & --- & --- & --- & --- \\
Video-R1~\cite{video-r1} & --- & --- & 96.44 & --- & --- & --- & --- & --- & --- & --- \\
\midrule

\rowcolor{SectionGray} \multicolumn{11}{l}{\textit{Segmentation Methods}} \\
SAM3~\cite{sam3} & --- & --- & --- & --- & --- & --- & --- & --- & --- & 30.13 \\
\midrule

OmniVL-Guard~\cite{omnivl-guard} & 96.20 & 93.12 & 98.58 & 75.52 & --- & --- & 54.26 & 63.78 & 59.22 & --- \\
\rowcolor{LightGreen}
\textbf{OmniVL-Guard Pro (Ours)} & \textbf{97.38} & \textbf{94.67} & \textbf{99.03} & \textbf{78.91} & \textbf{81.27} & \textbf{86.45} & \textbf{58.72} & \textbf{66.94} & \textbf{64.37} & \textbf{55.90} \\
\midrule

\textbf{$\Delta$ (vs. Best)} &
{\color{red} +1.18} & {\color{red} +1.55} & {\color{red} +0.22} & {\color{red} +3.39} & {\color{red} +26.96} & {\color{red} +30.71} &
{\color{red} +4.46} & {\color{red} +3.16} & {\color{red} +5.15} & {\color{red} +25.77} \\
\bottomrule
\end{tabular}
}
\vspace{-0.1cm}

\caption{Performance comparison on out-of-domain benchmarks.}
\vspace{-0.1cm}
\label{tab:outofdomain_comparison}
\resizebox{\textwidth}{!}{
\begin{tabular}{ll cc ccc cc cc cc c c}
\toprule
\multicolumn{2}{c}{} &
\multicolumn{2}{c}{\cellcolor{SectionGray}\textbf{MLLMs}} &
\multicolumn{3}{c}{\cellcolor{SectionGray}\textbf{Image-Text Methods}} &
\multicolumn{2}{c}{\cellcolor{SectionGray}\textbf{Text Methods}} &
\multicolumn{2}{c}{\cellcolor{SectionGray}\textbf{Image Methods}} &
\multicolumn{2}{c}{\cellcolor{SectionGray}\textbf{Video-Text Methods}} &
\multicolumn{1}{c}{\cellcolor{SectionGray}\textbf{Seg.}} &
\multicolumn{1}{c}{\cellcolor{LightGreen}\textbf{}} \\
\cmidrule(lr){3-4} \cmidrule(lr){5-7} \cmidrule(lr){8-9} \cmidrule(lr){10-11} \cmidrule(lr){12-13} \cmidrule(lr){14-14} \cmidrule(lr){15-15}
\textbf{Modality} & \textbf{Dataset} &
Qwen3VL & Llama3.2 &
HAMMER & FKA-Owl & AMD &
Bert & LUKE &
D$^3$ & Fake-VLM &
FakeSV-VLM & Video-R1 &
SAM3 &
\cellcolor{LightGreen}\textbf{Pro (Ours)} \\
\midrule
\textbf{Text} & ISOT &
88.74 & 81.91 & --- & --- & --- & 33.95 & 35.78 & --- & --- & --- & --- & --- & \cellcolor{LightGreen}\textbf{94.12} \\
\textbf{Image} & CASIA2.0 &
60.88 & 58.17 & --- & --- & --- & --- & --- & 49.20 & 51.22 & --- & --- & --- & \cellcolor{LightGreen}\textbf{67.48} \\
\textbf{Text-Image} & MMFakeBench &
57.40 & 54.14 & 61.83 & 62.32 & 59.11 & --- & --- & --- & --- & --- & --- & --- & \cellcolor{LightGreen}\textbf{81.74} \\
\textbf{Text-Video} & FakeTT &
59.84 & 56.28 & --- & --- & --- & --- & --- & --- & --- & 54.26 & 52.18 & --- & \cellcolor{LightGreen}\textbf{68.06} \\
\textbf{Segmentation} & IMD &
--- & --- & --- & --- & --- & --- & --- & --- & --- & --- & --- & 16.22 & \cellcolor{LightGreen}\textbf{42.93} \\
\bottomrule
\end{tabular}
}
\renewcommand{\arraystretch}{1.0}
\vspace{-0.5cm}
\end{table*}

\subsection{Stage II: Checker Construction via Rollout Distillation}
\label{sec:method_stage2}

After obtaining $\pi_{\mathrm{ans}}$, we construct a Checker to evaluate whether the complete tool reasoning trajectory produced by the Detector provides sufficient and consistent support for its final prediction, thereby avoiding the pseudo-success pattern where the ``answer is correct but the process is distorted.'' For each sample, the Checker takes the question $x$, the trajectory $\tau$, and the predicted answer $\hat{a}$ as input, and outputs fine-grained evaluation together with an overall score $s$.

We define the scoring space as three discrete levels: $s \in \{0,\ 0.5,\ 1\}$.
Specifically, $s=1$ indicates that the entire trajectory is internally consistent, the tool usage is reasonable, and the acquired key evidence is sufficient to support the final prediction; $s=0.5$ indicates that the trajectory is locally reasonable but still suffers from issues such as insufficient evidence or incomplete intermediate support; and $s=0$ indicates that the trajectory lacks effective support overall, the tool usage and intermediate reasoning fail to form a self-consistent evidence chain, or the final prediction is clearly inconsistent with the reasoning process. Details are provided in Appendix~\ref{sec:scoring_criteria}.

We adopt a discrete three-level score rather than a continuous dense score because the Checker is intended to provide process supervision signals with clear boundaries and stable judgments. In contrast, continuous scores often lack fine-grained discriminative criteria; for instance, the model cannot reliably provide a distinguishable boundary between 0.95 and 0.93, which would introduce additional noise.

For each question $x$ in $\mathcal{D}_{\mathrm{rl}}^{2}$, we use $\pi_{\mathrm{ans}}$ to perform rollout in the real tool environment, obtaining a complete response $y=(\tau,\hat{a})$. Based on these rollout responses, we further construct supervised samples for the Checker.

First, we perform preliminary filtering on the raw responses produced by $\pi_{\mathrm{ans}}$ rollout, removing samples with abnormal formats, incomplete trajectories, or parsing failures. We then construct an MLLM pool $\mathcal{M}=\{\text{GPT5.4},\ \text{Gemini3.1},\ \text{Seed2.0}\}$, and randomly select two MLLMs from it. Each selected MLLM is required to assess the trajectory according to the scoring criteria defined above, outputting both the fine-grained evaluation rationale and the discrete score $s$. Samples for which the two scores disagree are removed to complete cross-verification. Next, the remaining MLLM verifies the above evaluations and scores, filtering out samples with ambiguous evidence or unclear scoring boundaries. All samples that pass the filtering process jointly constitute the Checker training set $\mathcal{D}_{\mathrm{checker}}$. The construction details are provided in Appendix~\ref{sec:checker_prompts}.

We then concatenate the fine-grained evaluation and the discrete score into the target output sequence, and perform standard autoregressive SFT on Qwen3VL-8B using $\mathcal{D}_{\mathrm{checker}}$, obtaining the process discrimination model $\pi_{\mathrm{checker}}$.

\subsection{Stage III: Checker-Guided Process RL with ARSPO}

After obtaining the outcome-level policy $\pi_{\mathrm{ans}}$ and the process discrimination model $\pi_{\mathrm{checker}}$, we further conduct Checker-guided process-supervised RL, encouraging the model to generate more coherent, trustworthy, and concise tool reasoning trajectories while maintaining answer correctness.

For each task question $x$ in $\mathcal{D}_{\mathrm{rl}}^{2}$, the current policy performs inference in the real tool environment and produces $y=(\tau,\hat{a})$. We first determine whether the response is correct based on the final answer. For classification tasks, correctness means that the final prediction matches the annotation; for grounding and segmentation tasks, correctness means that the corresponding evaluation metric is higher than $0.7$. Only correct responses are further evaluated by $\pi_{\mathrm{checker}}$ to obtain a process score $s_{\mathrm{proc}}\in\{0,0.5,1\}$, while incorrect responses are not subjected to Checker-based process penalization, since they have already been penalized by the answer reward.

To avoid unconstrained tool invocation during inference, we further design a tool-use efficiency reward. Specifically, among all correct responses to the same question, the response with the fewest tool calls receives $R_{\mathrm{toolmin}}(y)=0.5$, while the others receive $R_{\mathrm{toolmin}}(y)=0$.

Based on the above design, we define the total reward as
\begin{equation}
R(y,x)=
\begin{cases}
s_{\mathrm{proc}}(y) \times
\bigl(R_{\mathrm{ans}}(y,x)+R_{\mathrm{fmt}}(y) \\
\quad +R_{\mathrm{toolmin}}(y)+R_{\mathrm{rep}}(y)\bigr),
& \text{if correct}, \\[3pt]
R_{\mathrm{ans}}(y,x)+R_{\mathrm{rep}}(y)+R_{\mathrm{fmt}}(y),
& \text{otherwise}.
\end{cases}
\end{equation}
where $R_{\mathrm{ans}}$, $R_{\mathrm{toolmin}}$, $R_{\mathrm{rep}}$, and $R_{\mathrm{fmt}}$ denote the answer reward, tool-use efficiency reward, repetition penalty, and format reward, respectively.

We still adopt ARSPO to update the policy, with the optimization objective defined as follows:
\begin{equation}
\begin{split}
\mathcal{J}_{\mathrm{arspo}}^{\mathrm{proc}}(\theta)
&= \sum_{k=1}^{K}\frac{|\mathcal{D}_{\mathrm{rl},k}^{2}|}{|\mathcal{D}_{\mathrm{rl}}^{2}|}
\; \mathbb{E}_{\substack{x\sim \mathcal{D}_{\mathrm{rl},k}^{2},\\ \{y_i\}\sim \pi_{\theta_{\mathrm{old}}}}} \\
&\qquad \times \Bigg[
\frac{l_{k,s}}{G}\sum_{i=1}^{G}
\frac{1}{|y_i|}
\sum_{t=1}^{|y_i|}
f_{i,t}\big(r_{i,t}(\theta)\big)\hat{A}_{i,k}
\Bigg],
\end{split}
\label{eq:arspo_checker}
\end{equation}
The token-level optimization in Eq.~\eqref{eq:arspo_checker} is only applied to model-generated tokens. After this stage, the model is jointly optimized in terms of answer correctness, process consistency, and tool-use efficiency, ultimately yielding the process-supervision-enhanced policy.

During RL rollout and evaluation, each generated tool call is executed online, and its observation is fed back before the next step; see Appendix~\ref{sec:inference_protocol}.

\begin{table*}[tb]
\centering
\setlength{\tabcolsep}{4.2pt}
\renewcommand{\arraystretch}{1.15}
\caption{Ablation study of the four training components. The rightmost column shows the average improvement ($\Delta$ AVG) over the SFT baseline.}
\label{tab:ablation_components}
\resizebox{\textwidth}{!}{
\begin{tabular}{cccc|cc ccc c|c}
\toprule
\multicolumn{2}{c}{\textbf{Stage I}} &
\multicolumn{1}{c}{\textbf{Stage II}} &
\multicolumn{1}{c}{\textbf{Stage III}} &
\multicolumn{2}{c}{\textbf{Classification}} &
\multicolumn{3}{c}{\textbf{Localization}} &
\multicolumn{1}{c}{\textbf{Seg.}} &
\multicolumn{1}{c}{} \\
\cmidrule(lr){1-2} \cmidrule(lr){3-3} \cmidrule(lr){4-4} \cmidrule(lr){5-6} \cmidrule(lr){7-9} \cmidrule(lr){10-10} \cmidrule(lr){11-11}
\textbf{SFT} & \textbf{ARSPO} & \textbf{Checker Train} & \textbf{Checker RL} &
\textbf{Bin. Cls.} & \textbf{RealFact} &
\textbf{Image} & \textbf{Text} & \textbf{Video} &
\textbf{Dice} &
\textbf{$\Delta$ AVG} \\
\midrule
\cmark & \nmark & \nmark & \nmark &
60.74 & 70.82 & 28.30 & 43.48 & 19.64 & 32.35 & -- \\
\cmark & \cmark & \nmark & \nmark &
88.48 & 81.36 & 52.41 & 63.92 & 50.35 & 48.64 & +21.64 \\
\cmark & \cmark & \nmark & \cmark &
84.27 & 77.84 & 46.83 & 60.71 & 40.08 & 43.27 & +16.28 \\
\rowcolor{LightGreen}
\cmark & \cmark & \cmark & \cmark &
\textbf{91.92} & \textbf{86.45} & \textbf{58.72} & \textbf{66.94} & \textbf{64.37} & \textbf{55.90} & \textbf{+28.16} \\
\bottomrule
\end{tabular}
}
\renewcommand{\arraystretch}{1.0}
\vspace{-0.5cm}
\end{table*}

\begin{figure*}[tb]
    \centering
    \includegraphics[width=\fulltextwidth]{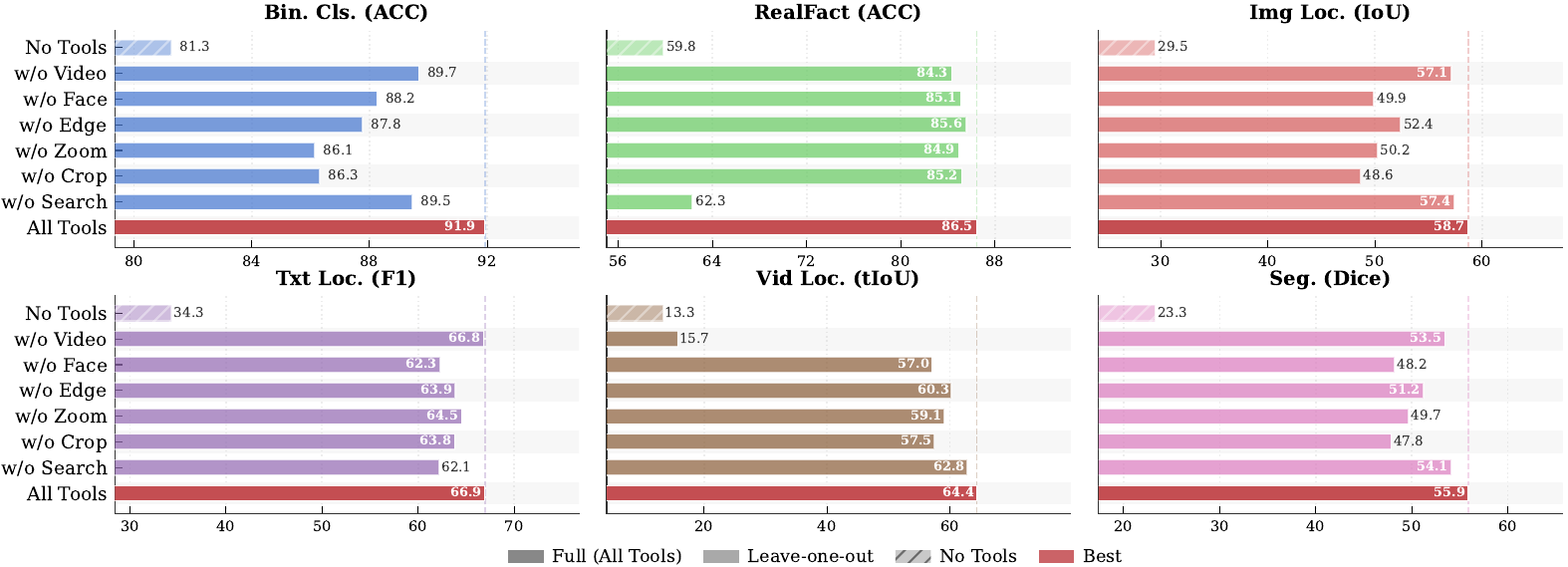}
    \caption{Tool ablation study. Each panel shows the performance of one metric across different tool configurations. The dashed line indicates the All Tools baseline.}
    \label{fig:tools_ablation}
\end{figure*}

\section{Experiments}
\label{sec:experiments}
Implementation details are provided in Appendix~\ref{sec:implementation}.

\textbf{Benchmarks and Baselines.}
We evaluate on both In-Domain and Out-of-Domain (OOD) benchmarks. The In-Domain setting uses the test split from Sec.~\ref{sec:in-domain-datasets}, while the OOD setting is designed to assess zero-shot generalization. The OOD suite includes ISOT~\cite{isot} for text, CASIA2.0~\cite{casia2.0} for image, MMFakeBench~\cite{mmfakebench} for image-text, FakeTT~\cite{fakett} for video-text, and IMD~\cite{imd} for segmentation. For baselines, general MLLMs (e.g., Qwen3VL-235B) are evaluated in a zero-shot manner. In contrast, domain-specific methods are trained and then directly evaluated on OOD benchmarks without further fine-tuning, in order to assess cross-domain robustness.

\subsection{Main Results}


\textbf{In-Domain Comparison.}
As shown in Table~\ref{tab:in-domain}, OmniVL-Guard Pro achieves the best in-domain performance across detection, localization, real-time verification, and segmentation. While maintaining strong binary classification across all modalities, it delivers larger gains on fine-grained tasks, improving RealFact verification and segmentation by $30.71\%$ and $25.77\%$ over the prior results, validating the benefits of tool-augmented evidence acquisition and Checker-guided process supervision.


\textbf{Out-of-Domain Zero-Shot Comparison.}
As shown in Table~\ref{tab:outofdomain_comparison}, OmniVL-Guard Pro achieves the strongest OOD zero-shot performance across unseen benchmarks, outperforming both general MLLMs and domain-specific models. It attains $81.74\%$ and $68.06\%$ accuracy on Text-Image and Text-Video tasks, respectively, and reaches $42.93\%$ Dice on IMD segmentation. These results demonstrate that self-evolving tool reasoning and CGARL improve robustness to unseen multimodal forgeries and generalize beyond classification to fine-grained segmentation.

\subsection{Ablation Study}

\textbf{Ablation of Training Components.} As shown in Table~\ref{tab:ablation_components}, SFT alone provides only limited tool-use capability, while adding ARSPO brings a strong average gain of $21.64\%$, confirming the importance of outcome-level RL. Directly using Qwen3VL-8B as the intermediate process supervisor without training a dedicated Checker is less effective, yielding only $16.28\%$ average improvement, which suggests that generic MLLM feedback can introduce noisy process signals. With the trained Checker and Checker-guided RL, the full model achieves the best performance across all tasks and the largest average gain of $28.16\%$, validating the necessity of task-adapted process supervision.

\textbf{Ablation of Tool Modules.} Figure~\ref{fig:tools_ablation} shows that the full tool set consistently achieves the best overall performance, while removing all tools causes large drops across all tasks, confirming the necessity of external tool interaction. In particular, search is crucial for RealFact verification, where removing it reduces accuracy from $86.5\%$ to $62.3\%$. For video localization, removing the video tool leads to the largest degradation, reducing tIoU from $64.4\%$ to $15.7\%$. These results indicate that different tools provide complementary evidence for different forensic subtasks. 

\textbf{More Discussion.} We further investigate the impact of the self-evolution iteration rounds $N$ on model performance (Appendix~\ref{sec:ablation_iterations}) and the necessity of the Weakly-Hinted Hard Trajectory Synthesis for covering long-tail hard samples (Appendix~\ref{sec:ablation_hard_samples}).

\section{Conclusion}
We present OmniVL-Guard Pro, a tool-augmented agent that extends unified vision-language forensics from closed-world prediction to open-world evidence-driven reasoning. Through Tree-Structured Self-Evolving Trajectory Generation and Checker-Guided Agentic Reinforcement Learning, our model learns to actively acquire external clues, perform fine-grained inspection, and form evidence-consistent judgments. OmniVL-Guard Pro achieves state-of-the-art performance across all tasks, with strong zero-shot generalization.

{
\small

\bibliographystyle{icml2026}
\bibliography{references}
}

\clearpage
\onecolumn

\begin{center}
{\Large\bfseries Appendix}
\end{center}
\vspace{1.2em}
{
\renewcommand{\arraystretch}{1.1}
\begin{list}{}{\setlength{\leftmargin}{2.2em}\setlength{\itemsep}{0.25em}}
\item[\textbf{A.}] \hyperref[sec:related_work]{Related Work} \dotfill \pageref{sec:related_work}
\item[\textbf{B.}] \hyperref[sec:realtime_event_benchmark]{Real-time Event Benchmark} \dotfill \pageref{sec:realtime_event_benchmark}
\item[] \quad \hyperref[sec:construction_process]{B.1\ Construction Process} \dotfill \pageref{sec:construction_process}
\item[] \quad \hyperref[sec:prompts]{B.2\ Prompt Templates} \dotfill \pageref{sec:prompts}
\item[] \quad \hyperref[sec:human_verification]{B.3\ Human Verification} \dotfill \pageref{sec:human_verification}
\item[] \quad \hyperref[sec:examples]{B.4\ RealFact Benchmark Examples} \dotfill \pageref{sec:examples}

\item[\textbf{C.}] \hyperref[sec:seed_guidance_details]{Details of Tool Trajectory Seed Guidance} \dotfill \pageref{sec:seed_guidance_details}
\item[] \quad \hyperref[sec:guider_criteria]{C.1\ Guider Judgment in Seed Trajectory Construction} \dotfill \pageref{sec:guider_criteria}
\item[] \quad \hyperref[sec:quality_checking]{C.2\ Quality Checking Process} \dotfill \pageref{sec:quality_checking}
\item[] \quad \hyperref[sec:guider_weakly_hinted]{C.3\ Guider in Weakly-Hinted Hard Trajectory Synthesis} \dotfill \pageref{sec:guider_weakly_hinted}
\item[] \quad \hyperref[sec:refiner]{C.4\ Refiner for Hint Trace Elimination} \dotfill \pageref{sec:refiner}

\item[\textbf{D.}] \hyperref[sec:implementation]{Implementation Details} \dotfill \pageref{sec:implementation}
\item[] \quad \hyperref[sec:reward_functions]{D.1\ Reward Function Configuration} \dotfill \pageref{sec:reward_functions}
\item[] \quad \hyperref[sec:advantage_function]{D.2\ Advantage Weighting Function} \dotfill \pageref{sec:advantage_function}
\item[] \quad \hyperref[sec:dca_coefficients]{D.3\ Hyperparameter Settings in ARSPO} \dotfill \pageref{sec:dca_coefficients}

\item[\textbf{E.}] \hyperref[sec:inference_protocol]{Inference Protocol for Tool Interaction} \dotfill \pageref{sec:inference_protocol}

\item[\textbf{F.}] \hyperref[sec:checker_details]{Details of Checker Construction} \dotfill \pageref{sec:checker_details}
\item[] \quad \hyperref[sec:scoring_criteria]{F.1\ Scoring Criteria for Process Quality} \dotfill \pageref{sec:scoring_criteria}
\item[] \quad \hyperref[sec:checker_prompts]{F.2\ MLLM Prompts for Checker Construction} \dotfill \pageref{sec:checker_prompts}

\item[\textbf{G.}] \hyperref[sec:ablation_iterations]{Analysis of Self-Evolution Iteration Rounds} \dotfill \pageref{sec:ablation_iterations}
\item[\textbf{H.}] \hyperref[sec:ablation_hard_samples]{Ablation Study on Weakly-Hinted Hard Trajectory Synthesis} \dotfill \pageref{sec:ablation_hard_samples}

\item[\textbf{I.}] \hyperref[sec:limitations]{Limitations} \dotfill \pageref{sec:limitations}
\item[] \quad \hyperref[sec:open_source]{I.1\ Computational Cost and Open-Source Commitment} \dotfill \pageref{sec:open_source}
\item[] \quad \hyperref[sec:safeguards]{I.2\ Safeguards Against Misuse} \dotfill \pageref{sec:safeguards}

\item[\textbf{J.}] \hyperref[sec:case_study]{Case Study} \dotfill \pageref{sec:case_study}
\end{list}
}
\clearpage

\appendix

\section{Related Work}
\label{sec:related_work}

\textbf{Forgery Detection and Grounding.}
With the rapid evolution of generative techniques, detection methodologies have progressed from unimodal paradigms---targeting specific modalities such as images, text, or videos---to bimodal synergies (e.g., image--text, video--text). More recently, OmniVL-Guard establishes a unified framework that simultaneously handles detection and grounding across image, video, and text modalities within a single model. Nevertheless, all these methods operate under a closed-world paradigm that implicitly assumes verification can be completed by the model itself, leaving them vulnerable to manipulations requiring external evidence or fine-grained perceptual scrutiny.

\textbf{Tool-Augmented Language Agents.}
The integration of tool use into large language models has given rise to a new class of agents that interact with external environments to accomplish complex tasks. Foundational frameworks such as ReAct and Toolformer demonstrate that LLMs can effectively interleave reasoning with tool invocation. This paradigm has been extended to multimodal settings, where agents leverage specialized tools for visual question answering, image editing, and embodied reasoning. Despite these advances, the application of tool-augmented agents to forensics remains largely unexplored. Forgery detection, grounding, and segmentation present unique challenges: the agent must decide not only when and which tools to invoke, but also how to integrate heterogeneous evidence from diverse sources into a coherent and verifiable judgment.

\textbf{OmniVL-Guard Pro.}
Our work extends OmniVL-Guard from a closed-world unified forensic model to an open-world evidence-driven agent. Unlike prior methods that rely solely on parametric knowledge, OmniVL-Guard Pro actively retrieves real-time evidence through web search, performs fine-grained local inspection via cropping and zooming tools, and produces pixel-level manipulation masks using SAM3. To train this agent, we propose Tree-Structured Self-Evolving Trajectory Generation, which addresses the efficiency--bias dilemma inherent in constructing tool-reasoning data: direct MLLM exploration yields low-quality trajectories, while injecting ground-truth cues introduces hindsight bias. Furthermore, we introduce Checker-Guided Agentic Reinforcement Learning (CGARL) to provide process-level supervision that penalizes the pseudo-success pattern where the answer is accidentally correct but the reasoning process is distorted, a failure mode that answer-level rewards alone cannot detect.

\section{Real-time Event Benchmark}
\label{sec:realtime_event_benchmark}
\subsection{Construction Process}
\label{sec:construction_process}
To evaluate whether models possess real-time event verification capabilities beyond parametric memory, we construct a real-time event benchmark \textbf{RealFact} for open-world fact verification. The benchmark covers the period from January 1, 2025 to March 31, 2026, containing 6,208 positive samples and 6,208 negative samples across diverse domains, including politics, economy, technology, society, sports, and more. The construction process is detailed as follows:

\eventsub{Data Collection}
We first collect candidate real-time events from publicly accessible information sources to build an initial event pool. The sources include mainstream news media, official announcements, as well as public webpages such as sports announcements and policy release pages. Each candidate event is required to satisfy three criteria simultaneously: first, the event time must be explicit and fall within the target time window; second, publicly accessible external textual evidence must be available and cross-supported by multiple independent sources; third, the event must contain verifiable key elements, such as persons, organizations, locations, time, actions, or outcomes. Based on these criteria, we clean and deduplicate the raw candidate events, removing samples with ambiguous timestamps, single-source evidence, or insufficient verifiable support. The retained events constitute the base event set of the benchmark, covering politics, economy, technology, society, sports, disasters, international news, and other domains.

\eventsub{Evidence-grounded Sample Construction}
After obtaining the base event set, we employ an MLLM to extract key information from each event, forming structured information that includes the event time, subject, location, key action, which is subsequently verified by another MLLM. Based on the structured information, we construct benchmark samples. For positive samples, the textual claim is consistent with the canonical event representation and can be directly supported by external evidence. For negative samples, we construct deceptive claims around real events, rather than using fabricated statements that are obviously incorrect or semantically incoherent. Specifically, these negative samples are created by modifying the event time, replacing the event subject, altering location information, changing the key action or outcome, or incorrectly stitching together multiple real events.

\eventsub{Annotation and Verification}
To ensure the quality and difficulty of the benchmark, we further conduct multi-stage filtering on the constructed samples. First, we use GPT-5.4 to check whether samples contain explicit cues that directly leak the answer, and remove samples with obvious hinting information. Next, we use Qwen3VL-235B, whose knowledge base is up to early 2025, to reason over all samples without providing external evidence, filtering out simple samples that can be judged solely from parametric memory. Then, we provide Qwen3VL-235B with the corresponding external evidence and perform inference again, further removing ambiguous samples for which a clear conclusion still cannot be obtained even with evidence. Only samples that pass all the above checks are included in the final benchmark. Finally, we conduct human verification on a randomly sampled subset to validate the quality of the automated construction and filtering pipeline.

\subsection{Prompt Templates}
\label{sec:prompts}

This section provides the detailed prompt templates used in the benchmark construction process.

\begin{tcolorbox}[title=Prompt 1: Key Information Extraction, colback=gray!5, colframe=gray!50]
\textbf{System Instruction:}
You are an expert information extraction assistant. Your task is to extract structured information from news articles about real-world events.

\textbf{Your Task:}
Extract the following information from the provided news article:
\begin{enumerate}
    \item \textit{Event Time:} The specific date or time when the event occurred.
    \item \textit{Subject:} The person, organization, or entity involved in the event.
    \item \textit{Location:} The place where the event took place.
    \item \textit{Key Action:} The main action or event that occurred.
    \item \textit{Outcome:} The result or consequence of the event.
\end{enumerate}

\textbf{Input Data:}
\textbf{News Article:} \{article\_text\}

\textbf{Output Format:}
Return strictly in JSON format:
\{
    ``time'': ``...'',
    ``subject'': ``...'',
    ``location'': ``...'',
    ``action'': ``...'',
    ``outcome'': ``...''
\}
\end{tcolorbox}

\begin{tcolorbox}[title=Prompt 2: Information Verification, colback=gray!5, colframe=gray!50]
\textbf{System Instruction:}
You are an expert fact-checking assistant. Your task is to verify whether extracted information accurately reflects the original news article.

\textbf{Your Task:}
Verify each field of the extracted information against the original article:
\begin{enumerate}
    \item Is the event time correctly extracted?
    \item Is the subject correctly identified?
    \item Is the location correctly identified?
    \item Is the key action correctly described?
    \item Is the outcome correctly stated?
\end{enumerate}

\textbf{Input Data:}
\textbf{News Article:} \{article\_text\}
\textbf{Extracted Information:} \{extracted\_json\}

\textbf{Output Format:}
For each field, respond with ``Correct'' or ``Incorrect'' and provide a brief explanation if incorrect.
\end{tcolorbox}

\begin{tcolorbox}[title=Prompt 3: Leakage Check (GPT-5.4), colback=gray!5, colframe=gray!50]
\textbf{System Instruction:}
You are an expert quality assurance assistant. Your task is to check whether benchmark samples contain explicit cues that directly leak the answer.

\textbf{Your Task:}
Analyze the claim and evidence to detect any leakage:
\begin{enumerate}
    \item Does the claim contain obvious factual errors that can be detected without verification?
    \item Does the claim contain self-contradictory information?
    \item Does the claim contain explicit hints about its truthfulness?
\end{enumerate}

\textbf{Input Data:}
\textbf{Claim:} \{claim\_text\}
\textbf{Evidence:} \{evidence\_text\}

\textbf{Output Format:}
Respond with ``No Leakage'' if the claim does not contain obvious cues, or ``Leakage Detected'' with explanation.
\end{tcolorbox}

\begin{tcolorbox}[title=Prompt 4: Parametric Memory Reasoning (Qwen3VL-235B), colback=gray!5, colframe=gray!50]
\textbf{System Instruction:}
You are a knowledgeable assistant. Based solely on your internal knowledge (without any external evidence), determine whether the given claim is true or false.

\textbf{Your Task:}
Analyze the claim and provide your judgment:
\begin{enumerate}
    \item Consider the factual accuracy of the claim.
    \item Consider the temporal consistency (does the time match known events?).
    \item Consider the plausibility of the described event.
\end{enumerate}

\textbf{Input Data:}
\textbf{Claim:} \{claim\_text\}

\textbf{Output Format:}
Respond with:
\begin{itemize}
    \item ``True'' if you believe the claim is factually correct.
    \item ``False'' if you believe the claim is factually incorrect.
    \item ``Uncertain'' if you cannot determine the truthfulness from your knowledge.
\end{itemize}
\end{tcolorbox}

\begin{tcolorbox}[title=Prompt 5: Evidence-Grounded Reasoning (Qwen3VL-235B), colback=gray!5, colframe=gray!50]
\textbf{System Instruction:}
You are an expert fact-verification assistant. Based on the provided external evidence, determine whether the given claim is true or false.

\textbf{Your Task:}
Analyze the claim against the evidence:
\begin{enumerate}
    \item Does the evidence support the claim?
    \item Is there any contradiction between the claim and the evidence?
    \item Is the evidence sufficient to make a definitive judgment?
\end{enumerate}

\textbf{Input Data:}
\textbf{Claim:} \{claim\_text\}
\textbf{Evidence:} \{evidence\_text\}

\textbf{Output Format:}
Respond with:
\begin{itemize}
    \item ``Supported'' if the evidence confirms the claim is true.
    \item ``Refuted'' if the evidence contradicts the claim.
    \item ``Insufficient'' if the evidence is not enough to make a judgment.
\end{itemize}
\end{tcolorbox}

\subsection{Human Verification}
\label{sec:human_verification}

To validate the quality of the automated construction and filtering pipeline, we conduct human verification on a randomly sampled subset of the benchmark. This subsection details the verification process, criteria, and results.

\eventsub{Verification Criteria}
\begin{enumerate}
    \item \textbf{Claim Authenticity:} Does the claim accurately reflect the original event? For positive samples, the claim should be factually consistent with the event. For negative samples, the introduced manipulation should be realistic and non-trivial.
    \item \textbf{Evidence Sufficiency:} Is the provided evidence sufficient to verify or refute the claim? The evidence should be directly relevant and conclusive.
    \item \textbf{Temporal Consistency:} Is the event timestamp accurate and consistent with the claim? The temporal information should be verifiable through the evidence.
    \item \textbf{Manipulation Subtlety:} For negative samples, is the manipulation subtle enough to challenge models? Obvious or trivial manipulations are flagged for removal.
\end{enumerate}

\eventsub{Verification Results}
98.6\% of the sampled instances are accepted, indicating high quality of the automated pipeline.

These results demonstrate that our automated construction pipeline produces high-quality benchmark samples, with the vast majority meeting human standards for factuality, evidence sufficiency, and manipulation subtlety.

\subsection{RealFact Benchmark Examples}
\label{sec:examples}

\begin{figure}[htbp]
    \centering
    \includegraphics[width=0.9\linewidth]{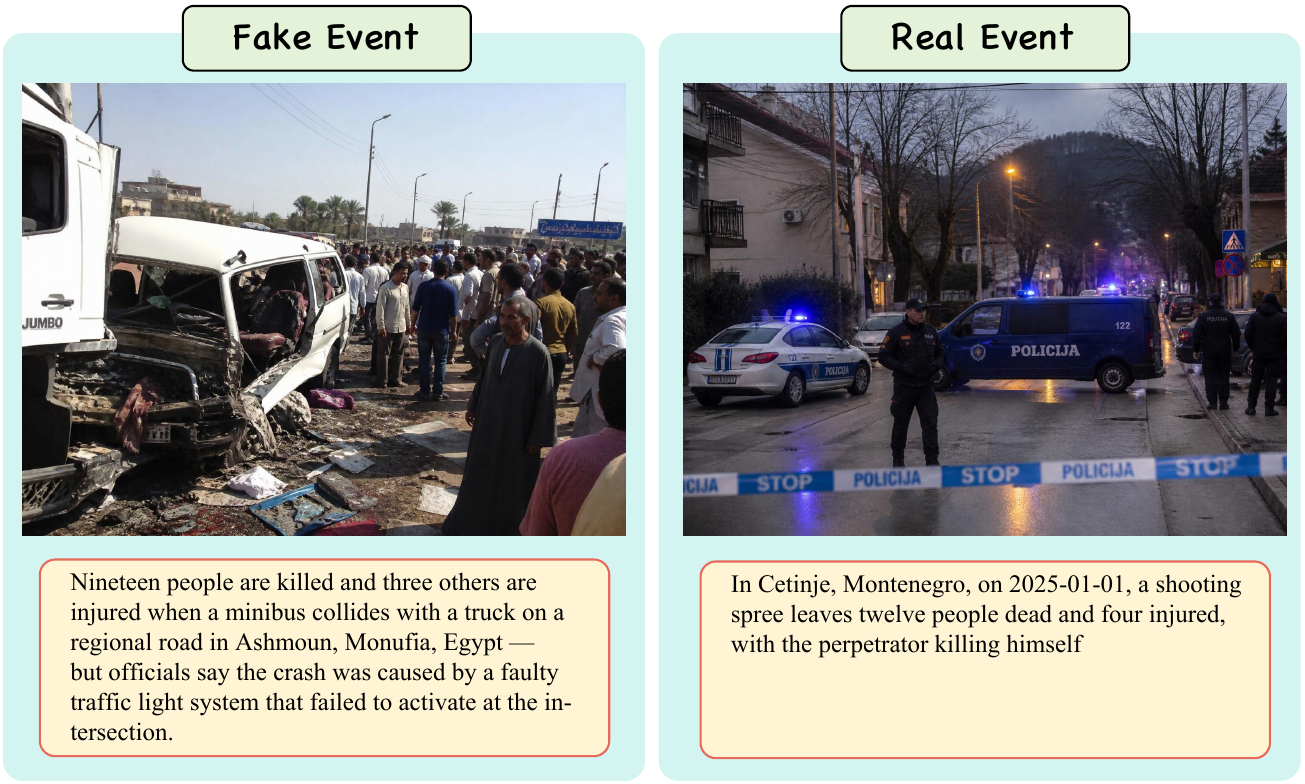}
    \caption{Representative examples from the RealFact benchmark. (a) A fake sample with manipulated content. (b) A real sample with verified factual content.}
    \label{fig:realfact_examples}
\end{figure}

Figure~\ref{fig:realfact_examples} presents two representative examples from the RealFact benchmark, illustrating a fake sample and a real sample.

\section{Details of Tool Trajectory Seed Guidance}
\label{sec:seed_guidance_details}

This section provides additional details for the Tree-Structured Self-Evolving Tool Trajectory Generation process described in Sec.~3. Specifically, we elaborate on four key components: (1) the \textit{Guider} judgment criteria used during seed trajectory construction (Sec.~\ref{sec:guider_criteria}); (2) the two-stage quality checking process for filtering high-quality seed trajectories (Sec.~\ref{sec:quality_checking}); (3) the \textit{Guider}'s dual role in weakly-hinted hard trajectory synthesis, where it both judges tool steps and provides weak hints to guide the Explorer (Sec.~\ref{sec:guider_weakly_hinted}); and (4) the \textit{Refiner} that eliminates explicit hint traces to produce autonomous-looking trajectories (Sec.~\ref{sec:refiner}).

\subsection{Guider Judgment in Seed Trajectory Construction}
\label{sec:guider_criteria}

During the tree-structured seed trajectory construction, the \textit{Guider} plays a critical role in evaluating each candidate tool step. At each step $t$, given the current history $\{(r_i,u_i,o_i)\}_{i=1}^{t-1}$ and the final answer $a^\star$, the \textit{Guider} assesses whether the candidate step $(r_t, u_t, o_t)$ is valid and helpful for solving the task. The \textit{Guider} judges the step based on whether the selected tool is appropriate for the current reasoning state, whether the tool parameters are reasonable, whether the returned observation provides useful information, and whether the step advances the reasoning toward the correct answer. If the step is judged invalid, the branch is immediately pruned; otherwise, the step is retained for further expansion.

The detailed prompt used for the \textit{Guider} judgment is provided below.

\begin{tcolorbox}[title=Prompt for Guider Judgment, colback=gray!5, colframe=gray!50]
\textbf{System Instruction:}
You are an expert Guider assisting an Explorer agent in constructing high-quality tool-use trajectories for vision-language forensics tasks. You have access to the ground truth answer $a^\star$.

\textbf{Your Task:}
Evaluate whether the candidate tool step is valid and helpful for solving the task. Consider the following criteria:
\begin{enumerate}
    \item \textbf{Tool Relevance:} Is the selected tool appropriate for the current task and reasoning state?
    \item \textbf{Parameter Quality:} Are the tool parameters (e.g., search query, crop box, zoom region) specific and reasonable?
    \item \textbf{Observation Utility:} Does the returned observation provide useful, non-redundant information?
    \item \textbf{Progress Toward Answer:} Does this step bring the reasoning closer to the ground truth answer $a^\star$?
\end{enumerate}

\textbf{Input Data:}
\textbf{Task Question:} \{question\}
\textbf{Ground Truth Answer:} $a^\star$
\textbf{Current History:} $\{(r_i, u_i, o_i)\}_{i=1}^{t-1}$
\textbf{Candidate Step:} $(r_t, u_t, o_t)$

\textbf{Output Format:}
Return strictly in JSON format:
\{
    ``judgment'': ``Valid'' or ``Invalid'',
    ``reason'': ``Brief explanation of the judgment.''
\}
\end{tcolorbox}

\subsection{Quality Checking Process}
\label{sec:quality_checking}

After the \textit{Explorer} completes the search and outputs the final prediction $\hat{a}$, we apply a two-stage quality checking process to ensure the trajectory meets the required standards.

\eventsub{Stage 1: Outcome-Based Filtering}
We first filter trajectories based on the correctness of the final prediction $\hat{a}$:
\begin{itemize}
    \item \textbf{Classification Tasks:} Trajectories with incorrect final answers are removed. Only trajectories where $\hat{a}$ matches the ground truth label $a^\star$ are retained.
    \item \textbf{Image Grounding Tasks:} We compute the Intersection over Union (IoU) between the predicted bounding box and the ground truth annotation. Only trajectories with IoU $\geq 0.7$ are retained.
    \item \textbf{Text Grounding Tasks:} We compute the F1 score between the predicted text span and the ground truth annotation. Only trajectories with F1 $\geq 0.7$ are retained.
    \item \textbf{Video Grounding Tasks:} We compute the temporal Intersection over Union (tIoU) between the predicted temporal segment and the ground truth annotation. Only trajectories with tIoU $\geq 0.7$ are retained.
    \item \textbf{Segmentation Tasks:} We compute the Dice score between the predicted mask and the ground truth mask. Only trajectories with Dice $\geq 0.7$ are retained.
    \item \textbf{Real-time Event Verification:} Trajectories with incorrect verification results (True/False) are removed.
\end{itemize}

\eventsub{Stage 2: Trajectory Quality Assessment}
For trajectories that pass Stage 1, we further assess the overall quality using the following criteria:
\begin{enumerate}
    \item \textbf{Trajectory Coherence:} The overall trajectory should be logically coherent, with each step building upon previous steps without abrupt jumps or contradictions.
    \item \textbf{Evidence Support:} The intermediate reasoning steps and tool usage should collectively support the final conclusion. The evidence gathered should directly verify or refute the claim, reveal manipulations, or localize relevant regions.
    \item \textbf{No Ineffective Steps:} The trajectory should not contain redundant tool invocations, empty observations, or reasoning steps that do not contribute to the final conclusion.
    \item \textbf{Completeness:} The trajectory should contain sufficient evidence to support the prediction, rather than arriving at the correct answer through lucky guessing.
\end{enumerate}

The detailed prompt used for trajectory quality assessment is provided below.

\begin{tcolorbox}[title=Prompt for Trajectory Quality Assessment, colback=gray!5, colframe=gray!50]
\textbf{System Instruction:}
You are an expert quality assessor for tool-use trajectories in vision-language forensics. Your task is to evaluate whether a trajectory meets the required quality standards.

\textbf{Your Task:}
Assess the trajectory based on the following criteria:
\begin{enumerate}
    \item \textbf{Coherence:} Is the trajectory logically coherent without abrupt jumps or contradictions?
    \item \textbf{Evidence Support:} Do the reasoning steps and tool usage collectively support the final conclusion?
    \item \textbf{Efficiency:} Does the trajectory contain ineffective or redundant steps?
    \item \textbf{Completeness:} Is there sufficient evidence to support the prediction?
\end{enumerate}

\textbf{Input Data:}
\textbf{Task Question:} \{question\}
\textbf{Ground Truth Answer:} $a^\star$
\textbf{Trajectory:} $\tau = \{(r_t, u_t, o_t)\}_{t=1}^{T} \cup \{\hat{a}\}$
\textbf{Final Prediction:} $\hat{a}$

\textbf{Output Format:}
Return strictly in JSON format:
\{
    ``quality'': ``Pass'' or ``Fail'',
    ``reason'': ``Brief explanation of the quality assessment.''
\}
\end{tcolorbox}

Only trajectories that pass both stages of the quality checking process are included in the seed set $\mathcal{D}_s^0$. This rigorous filtering ensures that the seed trajectories used for SFT are of high quality, providing a strong foundation for subsequent self-evolution and RL training.

\subsection{Guider in Weakly-Hinted Hard Trajectory Synthesis}
\label{sec:guider_weakly_hinted}

In the Weakly-Hinted Hard Trajectory Synthesis stage (Sec.~3.4), the \textit{Guider} plays a dual role compared to the Seed Guidance stage. Beyond judging whether the current tool step is valid, the \textit{Guider} also provides \textit{weak hints}---necessary parameters or local clues related to the current tool invocation---to narrow down the search space for the \textit{Explorer}. For example, in image scenarios, the \textit{Guider} provides a rough candidate region to assist cropping or zooming; in text scenarios, it may suggest keywords or entity names for search queries. These weak hints help the \textit{Explorer} overcome the difficulty of exploring hard samples, while keeping the hints sufficiently vague to avoid directly leaking the answer.

The detailed prompt used for the \textit{Guider} with weak hints is provided below.

\begin{tcolorbox}[title=Prompt for Guider with Weak Hints, colback=gray!5, colframe=gray!50]
\textbf{System Instruction:}
You are an expert Guider assisting an Explorer agent in constructing tool-use trajectories for hard samples in vision-language forensics. You have access to the ground truth answer $a^\star$. Your role is to evaluate tool steps and provide weak hints to guide the Explorer.

\textbf{Your Task:}
For each candidate tool step, perform two actions:
\begin{enumerate}
    \item \textbf{Judgment:} Evaluate whether the tool step is valid based on tool relevance, parameter quality, observation utility, and progress toward the answer.
    \item \textbf{Weak Hint:} If the step is invalid or suboptimal, provide a weak hint to guide the Explorer. The hint should be:
    \begin{itemize}
        \item Sufficient to narrow down the search space (e.g., a rough region for cropping, keywords for search).
        \item Vague enough to avoid directly revealing the answer $a^\star$.
        \item Focused on local clues rather than global conclusions.
    \end{itemize}
\end{enumerate}

\textbf{Input Data:}
\textbf{Task Question:} \{question\}
\textbf{Ground Truth Answer:} $a^\star$
\textbf{Current History:} $\{(r_i, u_i, o_i)\}_{i=1}^{t-1}$
\textbf{Candidate Step:} $(r_t, u_t, o_t)$

\textbf{Output Format:}
Return strictly in JSON format:
\{
    ``judgment'': ``Valid'' or ``Invalid'',
    ``hint'': ``Weak hint for the Explorer (empty if valid)'',
    ``reason'': ``Brief explanation of the judgment.''
\}
\end{tcolorbox}

\subsection{Refiner for Hint Trace Elimination}
\label{sec:refiner}

After obtaining the weakly-hinted candidate trajectories, we employ a \textit{Refiner} to eliminate explicit traces introduced by the weak hints. The \textit{Refiner} is a randomly selected model from the MLLM expert pool $\mathcal{M}$ (distinct from the Explorer and Guider), which rewrites the reasoning process $r_t$ at each step of the trajectory. The goal is to remove answer-aware hints from the reasoning, preventing the model from learning hint-dependent reasoning patterns that do not generalize to real inference scenarios where no hints are available.

Specifically, the \textit{Refiner} rewrites each reasoning step to:
\begin{enumerate}
    \item Remove explicit references to the hint (e.g., ``the Guider suggested region X'' $\rightarrow$ ``based on visual analysis, region X appears suspicious'').
    \item Rephrase the reasoning to sound as if it were generated autonomously by the Explorer.
    \item Preserve the factual content and logical flow of the original reasoning.
\end{enumerate}

The detailed prompt used for the \textit{Refiner} is provided below.

\begin{tcolorbox}[title=Prompt for Refiner (Hint Trace Elimination), colback=gray!5, colframe=gray!50]
\textbf{System Instruction:}
You are an expert Refiner responsible for rewriting reasoning processes in tool-use trajectories. Your task is to eliminate explicit traces of weak hints that were provided by a Guider, so that the final trajectory appears as if it were generated autonomously without any external guidance.

\textbf{Your Task:}
For each reasoning step $r_t$ in the trajectory, rewrite it to:
\begin{enumerate}
    \item Remove any explicit references to hints (e.g., ``the Guider suggested...'', ``based on the hint...'').
    \item Rephrase the reasoning to sound natural and autonomous, as if the Explorer discovered the information through its own analysis.
    \item Preserve the factual content, logical flow, and tool invocation decisions.
    \item Ensure the rewritten reasoning is consistent with the tool invocation $u_t$ and the returned observation $o_t$.
\end{enumerate}

\textbf{Input Data:}
\textbf{Task Question:} \{question\}
\textbf{Original Trajectory:} $\tau = \{(r_t, u_t, o_t)\}_{t=1}^{T} \cup \{\hat{a}\}$
\textbf{Weak Hints Used:} \{hints provided by the Guider at each step\}

\textbf{Output Format:}
Return the rewritten trajectory with hint traces eliminated:
$\tau' = \{(r'_t, u_t, o_t)\}_{t=1}^{T} \cup \{\hat{a}\}$
where $r'_t$ is the rewritten reasoning process for step $t$.
\end{tcolorbox}

All trajectories that pass the Refiner and the subsequent quality checking process (same as in Sec.~\ref{sec:quality_checking}) are included in the hard-sample set $\mathcal{D}_{sh}$.

\section{Implementation Details}
\label{sec:implementation}

We train our model on 32 NVIDIA H200 GPUs (141 GB) using Qwen3VL-8B as the backbone, and adopt AdamW throughout all training stages. In Stage I of CGARL, the SFT phase uses a batch size of 32, a learning rate of $6\times10^{-5}$, and 3 epochs, while the RL phase uses a batch size of 64 (i.e., 64 sampled trajectories per step), a learning rate of $1.2\times10^{-6}$, 2 epochs, and a KL penalty of 0.01. For the three fine-grained tasks—image tampering localization, text tampering localization, video temporal tampering localization and segmentation—the parameter $\alpha$ in the reward mapping function is consistently set to 3. The data scale ratio between $\mathcal{D}_{\mathrm{rl}}^{1}$ and $\mathcal{D}_{\mathrm{rl}}^{2}$ is approximately $7:3$.

\subsection{Reward Function Configuration}
\label{sec:reward_functions}

In the RL phase, the total reward $R$ for a generated response $y$ given question $x$ and ground truth $a^\star$ is composed of three distinct components: the task-specific performance score ($R_{\text{task}}$), the structural format reward ($R_{\text{fmt}}$), and a repetition penalty ($R_{\text{rep}}$). The final reward is calculated as:
\begin{equation}
    R(y, x) = R_{\text{task}}(y, x) + R_{\text{fmt}}(y) + R_{\text{rep}}(y)
\end{equation}

\noindent \textbf{1. Task-Specific Performance Reward ($R_{\text{task}}$)} \\
We employ distinct metrics for different forensic tasks. To encourage the model to strive for high-precision grounding rather than mediocre overlap, we apply an exponential mapping function to continuous metrics (IoU, F1, tIoU).

\begin{itemize}
    \item \textbf{Binary Forgery Classification:} For the binary classification task, we use a simple discrete reward based on correctness:
    \begin{equation}
        R_{\text{task}}^{\text{cls}} = \begin{cases}
        1, & \text{if } \hat{a} = a^\star, \\
        0, & \text{otherwise}.
        \end{cases}
    \end{equation}

    \item \textbf{Tampering Localization (Image, Text, Video):} For fine-grained localization tasks, we compute a raw metric $m \in [0, 1]$ specific to the modality: Intersection over Union (IoU) for image bounding boxes, F1-score for text indices, and Temporal IoU (tIoU) for video segments. The raw metric is mapped to a normalized reward using an exponential function with a scaling factor $\alpha = 3$:
    \begin{equation}
        R_{\text{task}}^{\text{loc}}(m) = \frac{e^{\alpha \cdot m} - 1}{e^{\alpha} - 1}
    \end{equation}
    This convex mapping ($\alpha=3$) amplifies the reward signal for high-quality responses (e.g., $m > 0.7$) while suppressing rewards for low-quality overlaps.

    \item \textbf{Forgery Segmentation:} For the segmentation task, we compute a composite score that combines box-level IoU and point prompt consistency. Specifically, we first convert the predicted bounding box into a binary mask $M_{\text{pred}}$ (1 inside the box, 0 outside) of the same size as the ground truth mask $M_{\text{gt}}$, and compute the Box IoU:
    \begin{equation}
        \text{IoU}_{\text{box}} = \frac{|M_{\text{pred}} \cap M_{\text{gt}}|}{|M_{\text{pred}} \cup M_{\text{gt}}|}
    \end{equation}
    We then evaluate point prompt consistency: for each predicted point, we check whether its label is consistent with the ground truth mask (i.e., positive points should fall inside $M_{\text{gt}}$ and negative points should fall outside). The point consistency score is computed as:
    \begin{equation}
        \text{score}_{\text{point}} = \frac{\text{number of consistent points}}{\text{total number of points}}
    \end{equation}
    The final segmentation score is a weighted combination:
    \begin{equation}
        m_{\text{seg}} = 0.7 \times \text{IoU}_{\text{box}} + 0.3 \times \text{score}_{\text{point}}
    \end{equation}
    clipped to $[0, 1]$. This composite score is then mapped through the same exponential function with $\alpha = 3$:
    \begin{equation}
        R_{\text{task}}^{\text{seg}}(m_{\text{seg}}) = \frac{e^{\alpha \cdot m_{\text{seg}}} - 1}{e^{\alpha} - 1}
    \end{equation}
\end{itemize}

\noindent \textbf{2. Format Compliance Reward ($R_{\text{fmt}}$)} \\
To enforce the tool trajectory structure, we employ a strict regular expression check. The model receives a positive reward if and only if the output strictly follows the required format:
\begin{equation}
    R_{\text{fmt}} =
    \begin{cases}
    0.2 & \text{if format is valid} \\
    0.0 & \text{otherwise}
    \end{cases}
\end{equation}

\noindent \textbf{3. Repetition Penalty ($R_{\text{rep}}$)} \\
To prevent degenerate generation loops, we implement an N-gram repetition penalty. We calculate the ratio of unique N-grams to total N-grams in the generated text. With a set hyperparameters of $N=3$ and a maximum penalty coefficient $\lambda_{\text{pen}} = -1.0$, the penalty is defined as:
\begin{equation}
    R_{\text{rep}} = \lambda_{\text{pen}} \times \left( 1 - \frac{|S_{\text{unique}}^{N}|}{N_{\text{total}}} \right)
\end{equation}
where $|S_{\text{unique}}^{N}|$ is the count of unique 3-grams and $N_{\text{total}}$ is the total number of 3-grams. This term introduces a negative reward proportional to the redundancy of the text.

\subsection{Advantage Weighting Function}
\label{sec:advantage_function}

The token-level weighting function $f_{i,t}(\cdot)$ in Eq.~\eqref{eq:arspo_tool} follows the SAPO design~\cite{sapo}. Specifically, for each response $y_i$ with normalized advantage $\hat{A}_{i,k}$, the weighting function is defined as:
\begin{equation}
f_{i,t}^{\text{sapo}}(r_{i,t}(\theta)) = \frac{4}{\tau_i} \sigma(\tau_i (r_{i,t}(\theta) - 1)), \quad \tau_i = \begin{cases} \tau_{\text{pos}}, & \widehat{A}_{i,k} > 0, \\ \tau_{\text{neg}}, & \widehat{A}_{i,k} \leq 0, \end{cases}
\end{equation}
where $r_{i,t}(\theta) = \frac{\pi_\theta(y_{i,t} \mid x, y_{i,<t})}{\pi_{\theta_{\mathrm{old}}}(y_{i,t} \mid x, y_{i,<t})}$ is the probability ratio, $\tau_{\text{neg}}$ and $\tau_{\text{pos}}$ denote the temperature parameters for positive and negative tokens, respectively, and $\sigma(x) = \frac{1}{1+e^{-x}}$ represents the sigmoid function.

\subsection{Hyperparameter settings in ARSPO}
\label{sec:dca_coefficients}

\begin{table}[H]
    \centering
    \caption{Hyperparameter settings for the Dynamic Coefficient Adjustment Algorithm in ARSPO.}
    \label{tab:hyperparams_dca}
    \vspace{0.2cm}
    \renewcommand{\arraystretch}{1.25}
    \setlength{\tabcolsep}{12pt}
    \begin{tabular}{l l c}
    \toprule
    \textbf{Symbol} & \textbf{Description} & \textbf{Value} \\
    \midrule
    \multicolumn{3}{l}{\textit{\textbf{Schedule \& Initialization}}} \\
    $T_{warm}$      & Warm-up phase duration (steps) & $800$ \\
    $T$             & Coefficient update interval (steps) & $100$ \\

    \midrule
    \multicolumn{3}{l}{\textit{\textbf{Adjustment Dynamics}}} \\
    $\alpha_{boost}$ & Coefficient boost factor & $1.1$ \\
    $\alpha_{decay}$ & Coefficient decay factor & $0.9$ \\

    \midrule
    \multicolumn{3}{l}{\textit{\textbf{Thresholds}}} \\
    $\epsilon_{mom}$    & Momentum protection threshold & $0.02$ \\
    $\epsilon_{rescue}$ & Regression rescue threshold & $0.10$ \\
    $\tau_{high}$       & High-performance threshold & $0.60$ \\
    \bottomrule
    \end{tabular}
\end{table}

\section{Inference Protocol for Tool Interaction}
\label{sec:inference_protocol}

At inference time, OmniVL-Guard Pro is executed as an interactive tool-augmented agent rather than a one-shot trajectory generator. Given an input question $x$, the policy first generates a reasoning segment $r_t$ and, when external clues or fine-grained inspection is needed, emits a structured tool invocation $u_t$ following the predefined action format. The environment parses $u_t$, executes the corresponding tool in $\mathcal{T}$, and returns the observation $o_t$, such as retrieved evidence, a cropped or zoomed image region, detected face boxes, extracted video frames, or a SAM3 segmentation mask. The observation $o_t$ is then appended to the context, and the policy continues generation conditioned on the updated interaction history $\{(r_i,u_i,o_i)\}_{i=1}^{t}$. This reasoning--action--observation loop is repeated until the policy emits a final answer $\hat{a}$ or reaches the maximum interaction budget.

Formally, for each test sample $x$, inference proceeds as
\[
r_t,u_t \sim \pi_\theta(\cdot \mid x,\{(r_i,u_i,o_i)\}_{i=1}^{t-1}), \qquad
o_t = \mathrm{Env}(u_t),
\]
where $\mathrm{Env}(\cdot)$ denotes the tool environment described in Sec.~\ref{sec:open_world_env}. The final prediction is produced only after the model has incorporated the observations returned by its previous tool calls:
\[
\hat{a} \sim \pi_\theta(\cdot \mid x,\{(r_i,u_i,o_i)\}_{i=1}^{T}).
\]
Therefore, the predicted trajectory is not merely a textual explanation generated after the fact; it records the actual sequence of tool invocations and observations used during inference.

We use the same tool interfaces, action schema, and observation format as in trajectory construction and RL training. To avoid unbounded exploration, inference is constrained by a maximum number of tool calls $T_{\max}$. If a generated tool invocation is malformed or refers to an unavailable tool, the environment returns an error observation and the model is allowed to continue, subject to the same interaction budget. Please refer to Appendix~\ref{sec:case_study} for examples.

\section{Details of Checker Construction}
\label{sec:checker_details}

This section provides additional details for the Checker construction process described in Sec.~4.2. Specifically, we elaborate on the scoring criteria for the discrete three-level score $s \in \{0, 0.5, 1\}$ and the prompts used for MLLM-based evaluation and verification.

\subsection{Scoring Criteria for Process Quality}
\label{sec:scoring_criteria}

The Checker assigns a discrete score $s \in \{0, 0.5, 1\}$ to evaluate the quality of the tool reasoning trajectory. The detailed criteria for each score level are as follows:

\noindent \textbf{Score $s=1$: Fully Consistent and Sufficient.}
A trajectory receives the highest score when it satisfies all of the following conditions:
\begin{itemize}
    \item \textbf{Internal Consistency:} The reasoning steps form a logically coherent chain without contradictions or abrupt jumps. Each step builds upon previous observations and maintains a clear narrative flow.
    \item \textbf{Reasonable Tool Usage:} The selected tools are appropriate for the current task and reasoning state. Tool parameters (e.g., search queries, crop boxes, zoom regions) are specific and well-justified. The model demonstrates understanding of when and how to invoke each tool.
    \item \textbf{Sufficient Evidence:} The acquired observations collectively provide sufficient evidence to support the final prediction. The evidence chain is complete, with no critical gaps or missing links that would undermine the conclusion.
    \item \textbf{Prediction-Process Alignment:} The final prediction $\hat{a}$ is clearly and directly supported by the accumulated evidence and reasoning. There is no contradiction between the reasoning process and the final answer.
\end{itemize}

\noindent \textbf{Score $s=0.5$: Partially Consistent with Limitations.}
A trajectory receives a moderate score when it exhibits local correctness but suffers from one or more of the following issues:
\begin{itemize}
    \item \textbf{Insufficient Evidence:} While the reasoning is locally sound, the trajectory fails to gather enough evidence to conclusively support the prediction. Some tool calls may return uninformative observations, or the model may prematurely conclude without exploring all relevant aspects.
    \item \textbf{Incomplete Intermediate Support:} The trajectory contains reasonable steps, but some intermediate conclusions are not fully supported by the preceding evidence. There may be minor logical gaps or assumptions that are not adequately validated.
    \item \textbf{Redundant but Non-Detracting Steps:} The trajectory includes some redundant tool calls or reasoning steps that do not contribute to the final conclusion, but do not actively mislead the reasoning process.
    \item \textbf{Weak Prediction Alignment:} The final prediction is plausible given the evidence, but the connection between the reasoning and the conclusion is not explicitly articulated or relies on implicit assumptions.
\end{itemize}

\noindent \textbf{Score $s=0$: Inconsistent or Insufficient.}
A trajectory receives the lowest score when it exhibits one or more of the following critical issues:
\begin{itemize}
    \item \textbf{Logical Inconsistency:} The reasoning chain contains contradictions, where later steps contradict earlier observations or conclusions. The trajectory lacks a coherent narrative structure.
    \item \textbf{Inappropriate Tool Usage:} The model selects inappropriate tools for the task, provides unreasonable parameters, or fails to interpret tool outputs correctly. The tool calls do not contribute meaningfully to the reasoning process.
    \item \textbf{Missing Evidence Chain:} The trajectory lacks a self-consistent evidence chain. Critical evidence is missing, and the final prediction is not supported by the available observations.
    \item \textbf{Prediction-Process Mismatch:} The final prediction $\hat{a}$ is clearly inconsistent with the reasoning process. The model may arrive at the correct answer through lucky guessing or flawed reasoning, which constitutes a pseudo-success pattern.
    \item \textbf{Format Violations:} The trajectory exhibits severe format issues, such as missing tool invocations, incomplete responses, or malformed output structures.
\end{itemize}

\subsection{MLLM Prompts for Checker Construction}
\label{sec:checker_prompts}

We provide the prompts used for the MLLM-based evaluation and verification in the Checker construction process.

\noindent \textbf{Prompt for Generating Score and Rationale.}
The following prompt is used by the two evaluation MLLMs to assess the trajectory and generate both the fine-grained evaluation rationale and the discrete score $s$:

\begin{tcolorbox}[title=Prompt for Evaluation MLLM (Score and Rationale Generation), colback=gray!5, colframe=gray!50]
\textbf{System Instruction:}
You are an expert evaluator for tool-use trajectories in vision-language forensics. Your task is to assess the quality of a reasoning trajectory that uses external tools to verify the authenticity of content. You must provide a detailed evaluation rationale and assign a discrete score based on the criteria below.

\textbf{Scoring Criteria:}

\textit{Score 1 (Fully Consistent and Sufficient):} The trajectory satisfies all of the following: (a) Internal Consistency---reasoning steps form a logically coherent chain without contradictions or abrupt jumps; (b) Reasonable Tool Usage---selected tools are appropriate for the task and reasoning state, and parameters are specific and well-justified; (c) Sufficient Evidence---acquired observations collectively provide sufficient evidence to support the final prediction, with no critical gaps; (d) Prediction-Process Alignment---the final prediction $\hat{a}$ is clearly and directly supported by the accumulated evidence and reasoning.

\textit{Score 0.5 (Partially Consistent with Limitations):} The trajectory exhibits local correctness but suffers from one or more of the following: (a) Insufficient Evidence---reasoning is locally sound but fails to gather enough evidence to conclusively support the prediction; (b) Incomplete Intermediate Support---some intermediate conclusions are not fully supported by the preceding evidence; (c) Redundant but Non-Detracting Steps---some redundant tool calls or reasoning steps that do not contribute to the final conclusion but do not actively mislead; (d) Weak Prediction Alignment---the final prediction is plausible but the connection between reasoning and conclusion is not explicitly articulated.

\textit{Score 0 (Inconsistent or Insufficient):} The trajectory exhibits one or more critical issues: (a) Logical Inconsistency---reasoning chain contains contradictions where later steps contradict earlier observations or conclusions; (b) Inappropriate Tool Usage---inappropriate tool selection, unreasonable parameters, or incorrect interpretation of tool outputs; (c) Missing Evidence Chain---critical evidence is missing and the final prediction is not supported by the available observations; (d) Prediction-Process Mismatch---the final prediction is clearly inconsistent with the reasoning process (pseudo-success pattern); (e) Format Violations---severe format issues such as missing tool invocations, incomplete responses, or malformed output structures.

\textbf{Input Data:}
\textbf{Task Question:} \{question\}
\textbf{Ground Truth Answer:} $a^\star$
\textbf{Trajectory:} $\tau = \{(r_t, u_t, o_t)\}_{t=1}^{T} \cup \{\hat{a}\}$
\textbf{Final Prediction:} $\hat{a}$

\textbf{Output Format:}
Return strictly in JSON format:
\{
    ``evaluation'': ``Detailed analysis of the trajectory quality. Specifically address: (1) whether the reasoning steps are internally consistent, (2) whether the tool usage is appropriate and well-parameterized, (3) whether the acquired evidence is sufficient to support the conclusion, and (4) whether the final prediction aligns with the reasoning process.'',
    ``score'': 0, 0.5, or 1
\}
\end{tcolorbox}

\noindent \textbf{Prompt for Verification.}
After an evaluation MLLM generates its evaluation and score, a third MLLM from the pool $\mathcal{M}$ independently verifies each evaluation. Specifically, for each evaluation-scorpair produced by an evaluation MLLM, the verification MLLM checks whether the evaluation is reasonable and the score is justified. The verification is performed one evaluation at a time, and the verification prompt is as follows:

\begin{tcolorbox}[title=Prompt for Verification MLLM (Single Evaluation Verification), colback=gray!5, colframe=gray!50]
\textbf{System Instruction:}
You are a quality assurance verifier for tool-use trajectory evaluations. An evaluator has assessed a trajectory and provided an evaluation rationale along with a score. Your task is to verify whether the evaluation is reasonable, accurate, and well-justified.

\textbf{Verification Criteria:}
\begin{itemize}
    \item Check whether the evaluation accurately reflects the trajectory quality according to the scoring criteria.
    \item Identify any factual errors in the evaluation (e.g., incorrect claims about tool usage or evidence).
    \item Assess whether the evidence cited in the evaluation is accurate and relevant to the trajectory.
    \item Determine if the score is justified given the evaluation rationale.
\end{itemize}

\textbf{Input Data:}
\textbf{Task Question:} \{question\}
\textbf{Ground Truth Answer:} $a^\star$
\textbf{Trajectory:} $\tau = \{(r_t, u_t, o_t)\}_{t=1}^{T} \cup \{\hat{a}\}$
\textbf{Final Prediction:} $\hat{a}$
\textbf{Evaluation:} \{evaluation from evaluation MLLM\}
\textbf{Score:} \{score from evaluation MLLM\}

\textbf{Output Format:}
Return strictly in JSON format:
\{
    ``verification'': ``Detailed analysis of whether the evaluation is reasonable and accurate. Highlight any factual errors or inconsistencies found.'',
    ``is\_valid'': true or false,
    ``reason'': ``Explanation for the verification decision. If invalid, specify which aspects of the evaluation contain errors.''
\}
\end{tcolorbox}

All samples that pass the verification process, where both evaluation MLLMs agree on the score and the verification MLLM confirms the validity, are included in the Checker training set $\mathcal{D}_{\mathrm{checker}}$.

\section{Analysis of Self-Evolution Iteration Rounds}
\label{sec:ablation_iterations}

To determine the optimal termination point for the Tree-Structured Self-Evolving Tool Trajectory Generation strategy (Sec.~\ref{sec:self-evolving}), we investigated the performance trajectory across iterations $n$. We hypothesized that the self-evolution process would eventually reach a saturation point where the teacher policy $\pi_n$ no longer provides superior tool-use signals compared to $\pi_{n-1}$, thereby yielding no further quality improvements in the generated dataset $\mathrm{FSTR}_{\mathrm{sft}}^n$.

Specifically, we compared the capabilities of the model trained after the fourth iteration against a model extended to a fifth iteration. We constructed a fifth-round dataset $\mathrm{FSTR}_{\mathrm{sft}}^5$ using $\pi_4$ and combined it with the hard-sample set $\mathcal{D}_{sh}$ to train the full OmniVL-Guard Pro framework via CGARL. We then compared this against our standard setting ($\mathrm{FSTR}_{\mathrm{sft}}^4 \cup \mathcal{D}_{sh}$).

\begin{table}[h]
\centering
\caption{Performance comparison between stopping at Iteration 4 ($\mathrm{FSTR}_{\mathrm{sft}}^4$) vs. Iteration 5 ($\mathrm{FSTR}_{\mathrm{sft}}^5$). The results indicate that performance plateaus at $n=4$, justifying the termination of the self-evolution loop to conserve computational resources.}
\label{tab:iteration_ablation}
\resizebox{0.9\linewidth}{!}{
\begin{tabular}{l|ccccc|cccc}
\toprule
\multirow{2}{*}{\textbf{Training Set}} & \multicolumn{5}{c|}{\textbf{Binary Classification (ACC)}} & \multicolumn{4}{c}{\textbf{Localization \& Segmentation}} \\
& Text & Image & Video & Text-Image & RealFact & Image (IoU) & Text (F1) & Video (tIoU) & Seg. (Dice) \\
\midrule
$\mathrm{FSTR}_{\mathrm{sft}}^4 \cup \mathcal{D}_{sh}$ (Ours) & 97.38 & 94.67 & 99.03 & 78.91 & 86.45 & 58.72 & 66.94 & 64.37 & 55.90 \\
$\mathrm{FSTR}_{\mathrm{sft}}^5 \cup \mathcal{D}_{sh}$ & 97.21 & 94.89 & 98.98 & 79.15 & 86.34 & 58.55 & 67.23 & 64.19 & 56.01 \\
\midrule
\textit{Diff ($\Delta$)} & \textit{+0.17} & \textit{-0.22} & \textit{+0.05} & \textit{-0.24} & \textit{+0.11} & \textit{+0.17} & \textit{-0.29} & \textit{+0.18} & \textit{-0.11} \\
\bottomrule
\end{tabular}
}
\end{table}

As illustrated in Table~\ref{tab:iteration_ablation}, the performance differences between the fourth and fifth iterations are statistically negligible. For instance, the Image Localization (IoU) shifted marginally from 58.72 to 58.55 ($\Delta=+0.17$), while Text Localization (F1) experienced a slight fluctuation from 66.94 to 67.23 ($\Delta=-0.29$). The results show a mix of minor improvements and degradations across tasks, with no consistent gains. This evidence suggests that the quality of the tool trajectory data generated by $\pi_4$ has reached a bottleneck, and further iterations do not yield distinguishable gains in downstream forensic tasks. Consequently, we terminate the self-evolution process at $n=4$ to strike an optimal balance between model performance and training efficiency.

\section{Ablation Study on Weakly-Hinted Hard Trajectory Synthesis}
\label{sec:ablation_hard_samples}

To validate the necessity of the \textbf{Weakly-Hinted Hard Trajectory Synthesis} stage (Sec.~3.4), we conducted an ablation study by excluding the hard-sample subset $\mathcal{D}_{sh}$ from the SFT cold-start dataset.

\noindent \textbf{Experimental Settings:}
\begin{itemize}
    \item \textbf{Full Setting (Ours):} The model is fine-tuned on the complete dataset $\mathrm{FSTR}_{\mathrm{sft}} = \mathcal{D}_s^4 \cup \mathcal{D}_{sh}$, followed by CGARL training.
    \item \textbf{w/o Hard Samples:} The model is fine-tuned only on the self-evolved dataset $\mathcal{D}_s^4$ (removing the synthesized hard samples $\mathcal{D}_{sh}$), followed by the same CGARL training.
\end{itemize}

The results are reported in Table~\ref{tab:ablation_hard_samples}. We observe that removing the hard samples leads to a consistent performance degradation across all tasks.

\begin{table}[h]
\centering
\caption{Ablation study on the necessity of Weakly-Hinted Hard Trajectory Synthesis. We report the performance comparison between using the full dataset versus removing the synthesized hard samples ($\mathcal{D}_{sh}$). The results show that $\mathcal{D}_{sh}$ is critical for maintaining high performance across both classification and localization tasks.}
\label{tab:ablation_hard_samples}
\resizebox{0.9\linewidth}{!}{
\begin{tabular}{l|ccccc|cccc}
\toprule
\multirow{2}{*}{\textbf{Method}} & \multicolumn{5}{c|}{\textbf{Binary Classification (ACC)}} & \multicolumn{4}{c}{\textbf{Localization \& Segmentation}} \\
 & Text & Image & Video & Text-Image & RealFact & Image (IoU) & Text (F1) & Video (tIoU) & Seg. (Dice) \\
\midrule
\textbf{Ours (Full $\mathrm{FSTR}_{\mathrm{sft}}$)} & \textbf{97.38} & \textbf{94.67} & \textbf{99.03} & \textbf{78.91} & \textbf{86.45} & \textbf{58.72} & \textbf{66.94} & \textbf{64.37} & \textbf{55.90} \\
w/o Hard Samples ($\mathcal{D}_{sh}$) & 93.45 & 88.72 & 92.18 & 71.56 & 82.23 & 52.84 & 57.31 & 56.42 & 48.23 \\
\midrule
\textit{Performance Drop ($\Delta$)} & \color{red}{-3.93} & \color{red}{-5.95} & \color{red}{-6.85} & \color{red}{-7.35} & \color{red}{-4.22} & \color{red}{-5.88} & \color{red}{-9.63} & \color{red}{-7.95} & \color{red}{-7.67} \\
\bottomrule
\end{tabular}
}
\end{table}

\textbf{Analysis of Results:}
\begin{itemize}
    \item \textbf{Impact on Fine-grained Localization and Segmentation:} The performance drop is significant in localization and segmentation tasks. Specifically, Text Localization (F1) drops by \textbf{9.63\%}, Video Temporal Localization (tIoU) drops by \textbf{7.95\%}, and Segmentation (Dice) drops by \textbf{7.67\%}. This empirical evidence suggests that the self-evolution loop alone tends to saturate on easier samples. The failure cases (hard samples) from the iterative process often contain subtle manipulations or complex multimodal inconsistencies that require the specific, high-quality tool-use trajectories provided by the Weakly-Hinted Hard Trajectory Synthesis to resolve.
    \item \textbf{Impact on Binary Classification:} The binary classification accuracy also decreases notably (e.g., \textbf{-7.35\%} on Text-Image). This indicates that the hard-sample data is indispensable not only for fine-grained grounding and segmentation but also for maintaining robust detection capabilities in complex scenarios.
\end{itemize}

In conclusion, the Weakly-Hinted Hard Trajectory Synthesis effectively complements the self-evolution pipeline by covering the long-tail distribution of difficult forgery types, preventing the model from overfitting to simple patterns.

\section{Limitations}
\label{sec:limitations}

\subsection{Computational Cost and Open-Source Commitment}
\label{sec:open_source}
Both the construction of the FSTR dataset and the multi-stage training of OmniVL-Guard Pro demand substantial computational resources, including large-scale tool-environment rollouts, multi-iteration self-evolution, and multi-stage RL optimization. Our goal is to establish a foundation model for unified vision-language forensics that supports diverse downstream tasks. To lower the barrier for the community and facilitate future research, we commit to fully open-sourcing the FSTR dataset, the trained OmniVL-Guard Pro checkpoints, and all associated code, enabling researchers to directly fine-tune or evaluate on their own tasks without reproducing the expensive training pipeline.

\subsection{Safeguards Against Misuse}
\label{sec:safeguards}
Given that the RealFact benchmark, FSTR dataset, and OmniVL-Guard Pro model could potentially be repurposed for malicious activities such as generating or evading forgery detection, we will implement a strict access control process. All assets will be released exclusively for academic research purposes, and each applicant will be required to provide institutional affiliation and a clear statement of intended use. Access will be granted only after a manual review, and any violation of the terms will result in immediate revocation.

\section{Case Study}
\label{sec:case_study}

\begin{figure}
    \centering
    \includegraphics[width=1\linewidth]{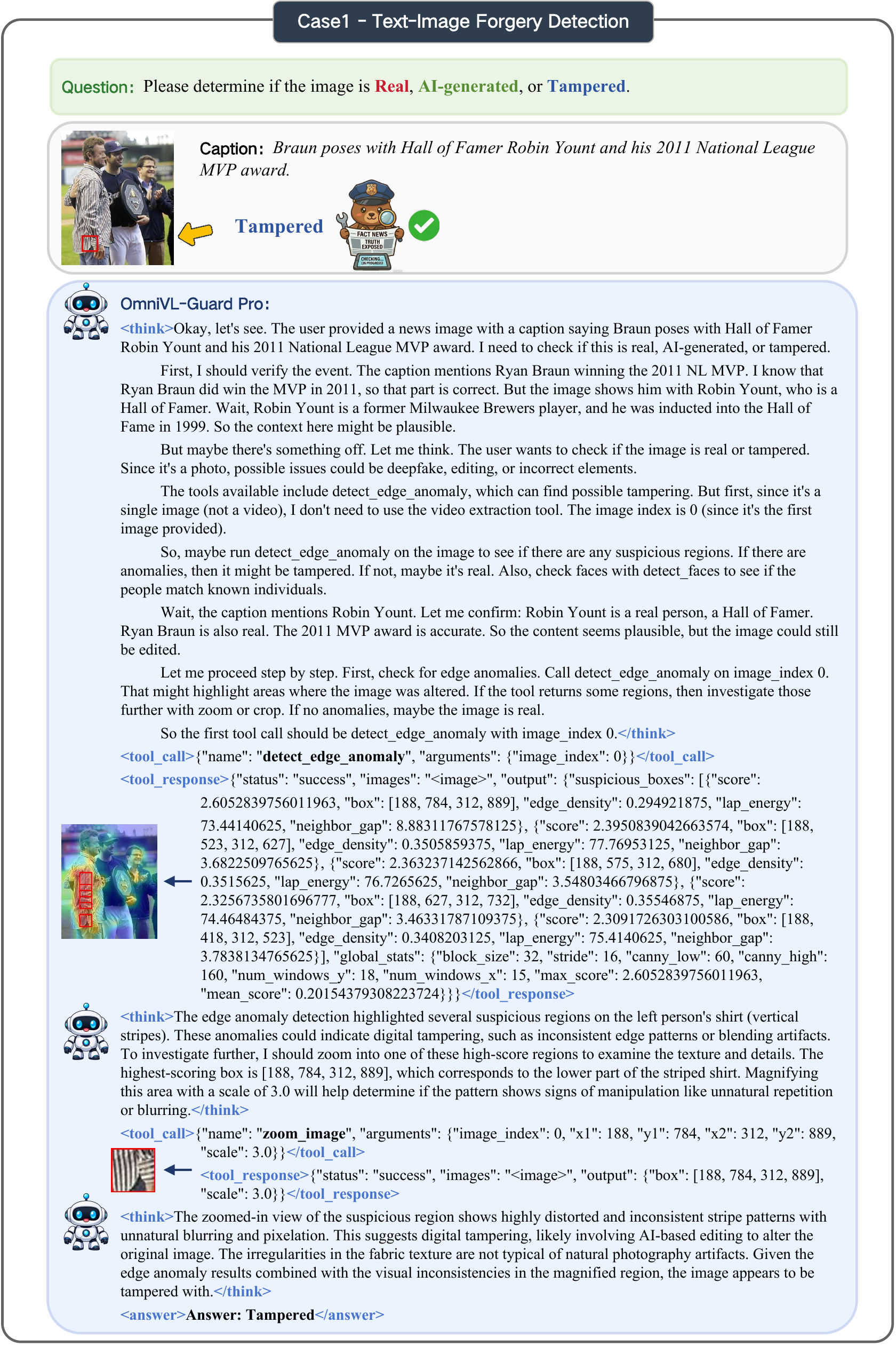}
\end{figure}

\begin{figure}
    \centering
    \includegraphics[width=1\linewidth]{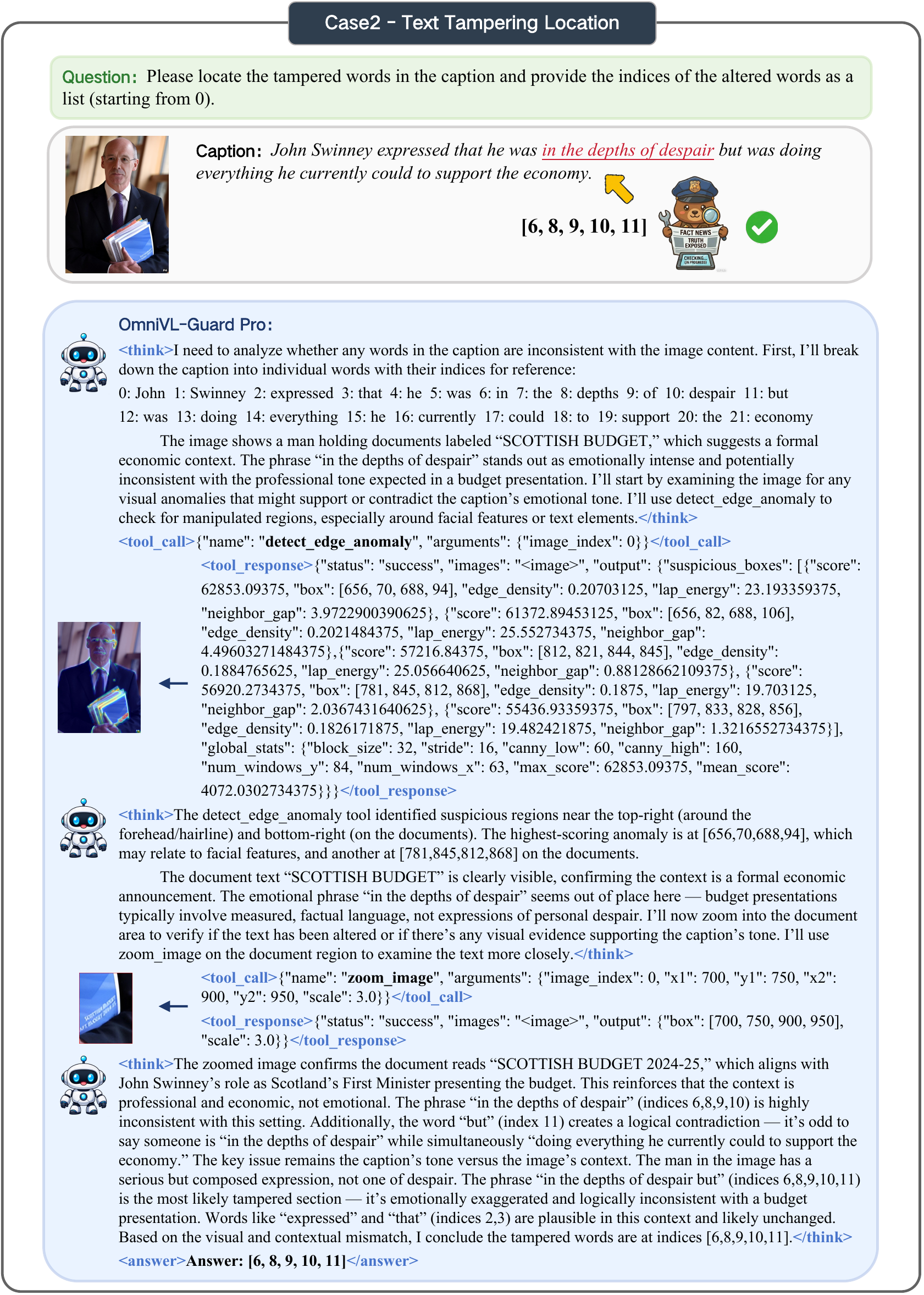}
\end{figure}

\begin{figure}
    \centering
    \includegraphics[width=1\linewidth]{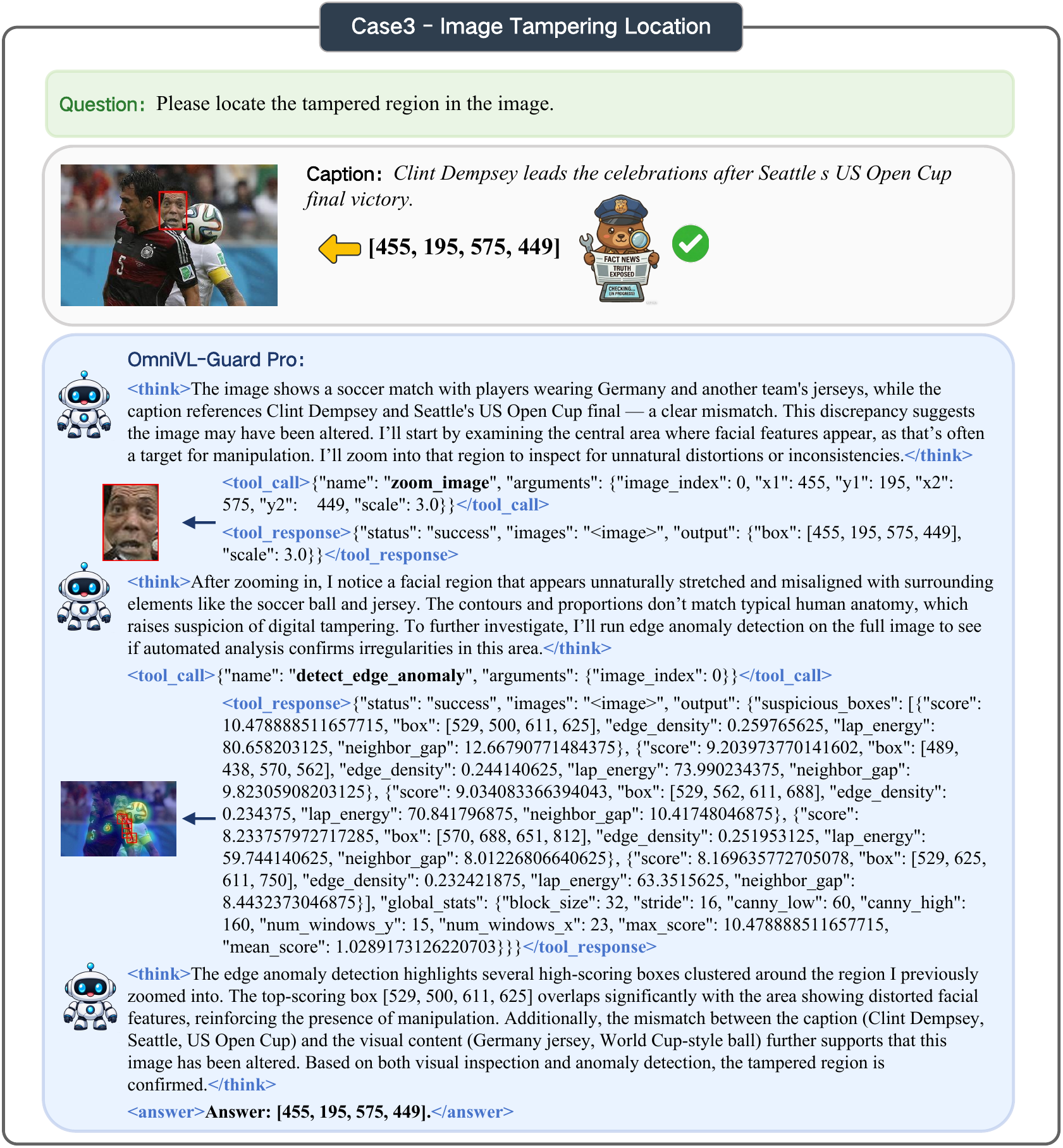}
\end{figure}

\begin{figure}
    \centering
    \includegraphics[width=0.7\linewidth]{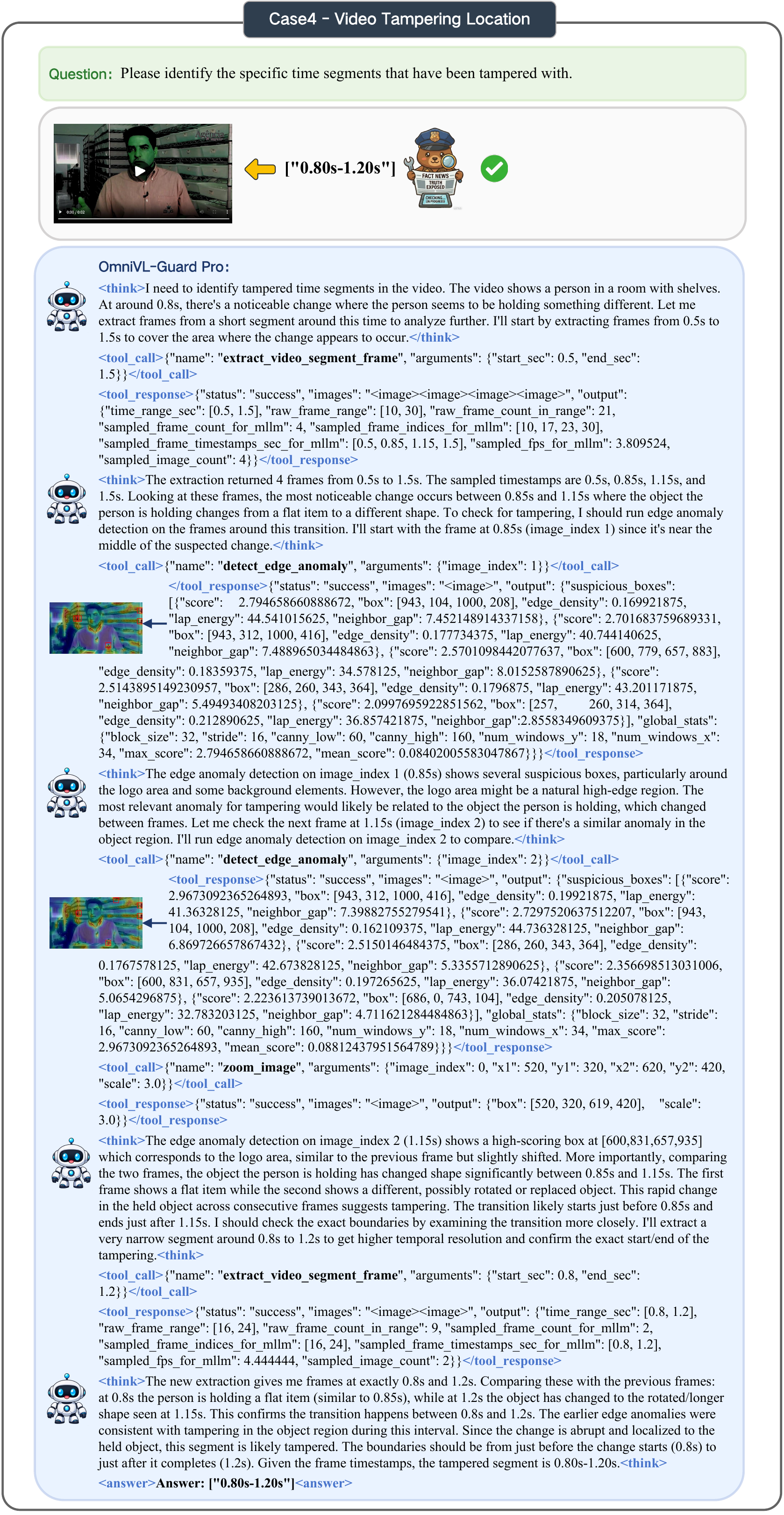}
\end{figure}

\begin{figure}
    \centering
    \includegraphics[width=0.8\linewidth]{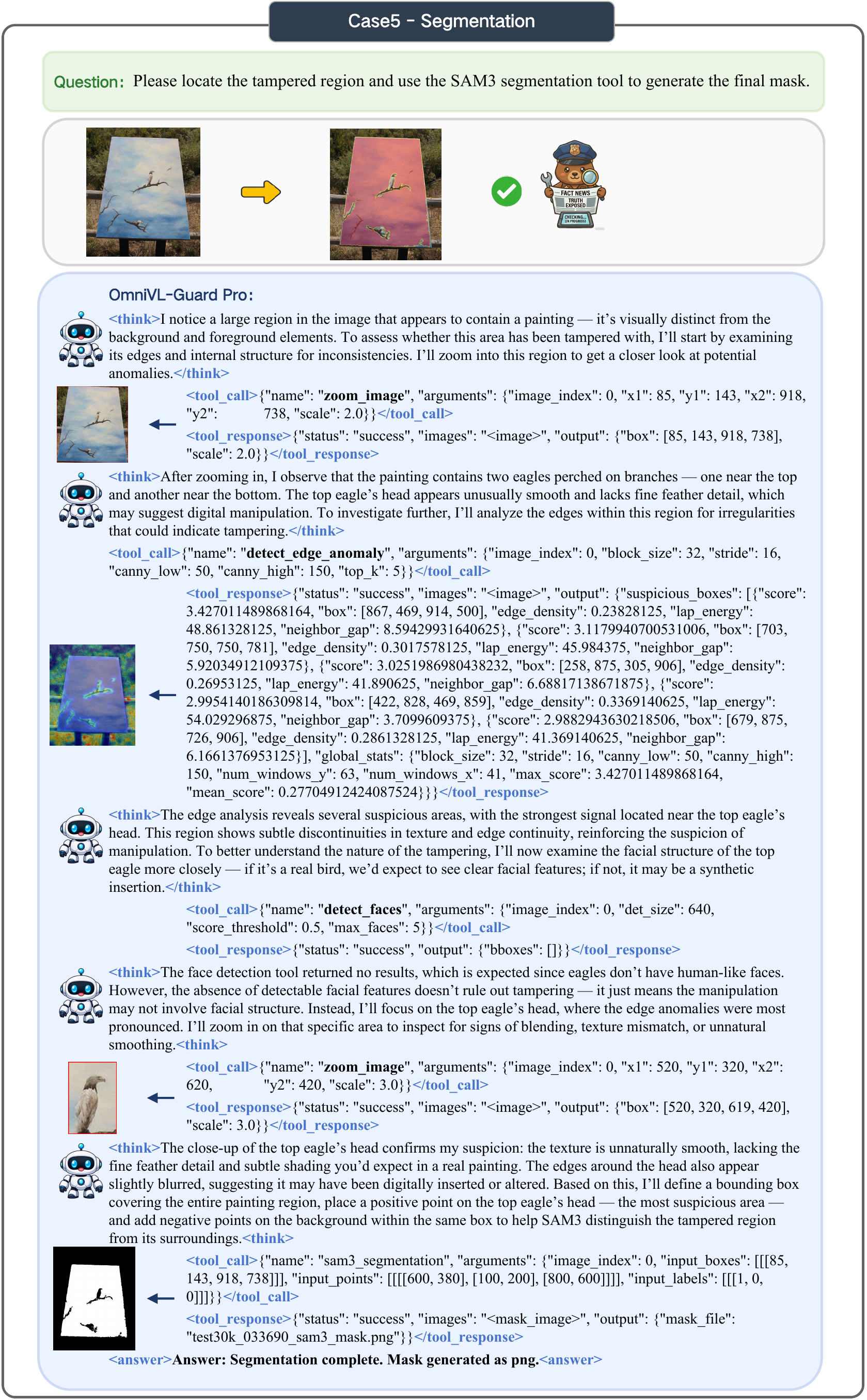}
\end{figure}


\makeatletter
\global\icml@noticeprintedtrue
\makeatother

\end{document}